\pdfoutput=1

\documentclass[11pt]{article}
\usepackage[dvipsnames]{xcolor}

\usepackage[preprint]{acl}

\usepackage{times}
\usepackage{latexsym}

\usepackage[T1]{fontenc}

\usepackage[utf8]{inputenc}

\usepackage{microtype}

\usepackage{inconsolata}

\usepackage{graphicx}
\usepackage{tabularx}
\usepackage{multirow}
\usepackage{amsfonts}
\usepackage{amssymb}
\usepackage{xspace}
\usepackage{lipsum}
\usepackage{adjustbox}
\usepackage{booktabs}
\usepackage{makecell}
\usepackage{url}
\usepackage{enumitem}
\setlist[enumerate]{topsep=0pt, partopsep=0pt, parsep=0pt, itemsep=0pt}
\setlist[itemize]{topsep=0pt, partopsep=0pt, parsep=0pt, itemsep=0pt}
\usepackage{listings}
\usepackage{xcolor}
\usepackage{inconsolata}
\usepackage[most]{tcolorbox}
\usepackage{caption}
\usepackage{float}
\usepackage{pgfplots}
\usepackage{subcaption}
\pgfplotsset{compat=1.17}
\usepackage{pgfplotstable}
\usepackage{amsthm}
\usepackage{pdfpages}
\newtheorem{definition}{Definition}

\newcommand{\unravel}{%
  \raisebox{-0.15ex}{%
    \includegraphics[height=1em]{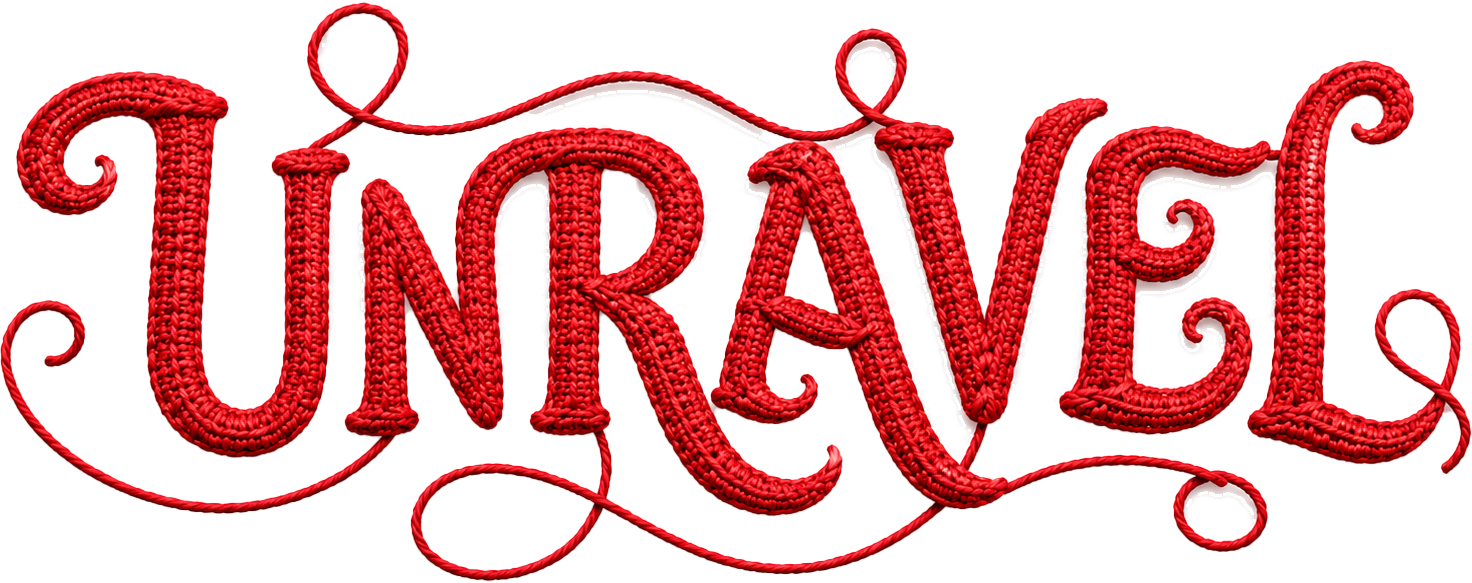}%
  }\xspace%
}

\newcommand{\drexel}{%
  \hspace{1pt}
  \begingroup\normalfont
  \includegraphics[height=1.3\fontcharht\font`\B]{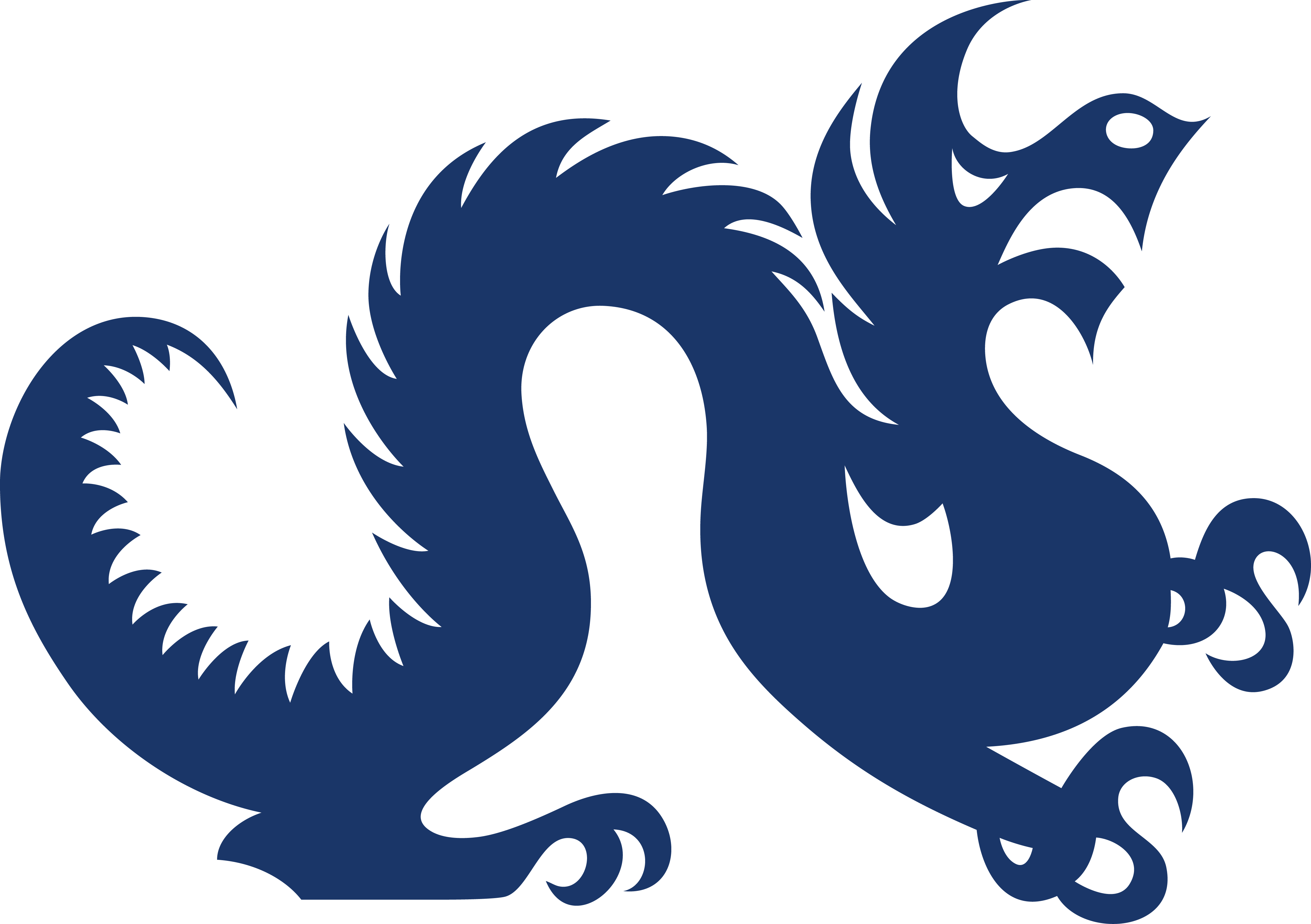}%
  \endgroup
  \hspace{1pt}
}
\newcommand{\aitwo}{%
  \hspace{1pt}%
  \begingroup\normalfont
  \includegraphics[height=1.3\fontcharht\font`\B]{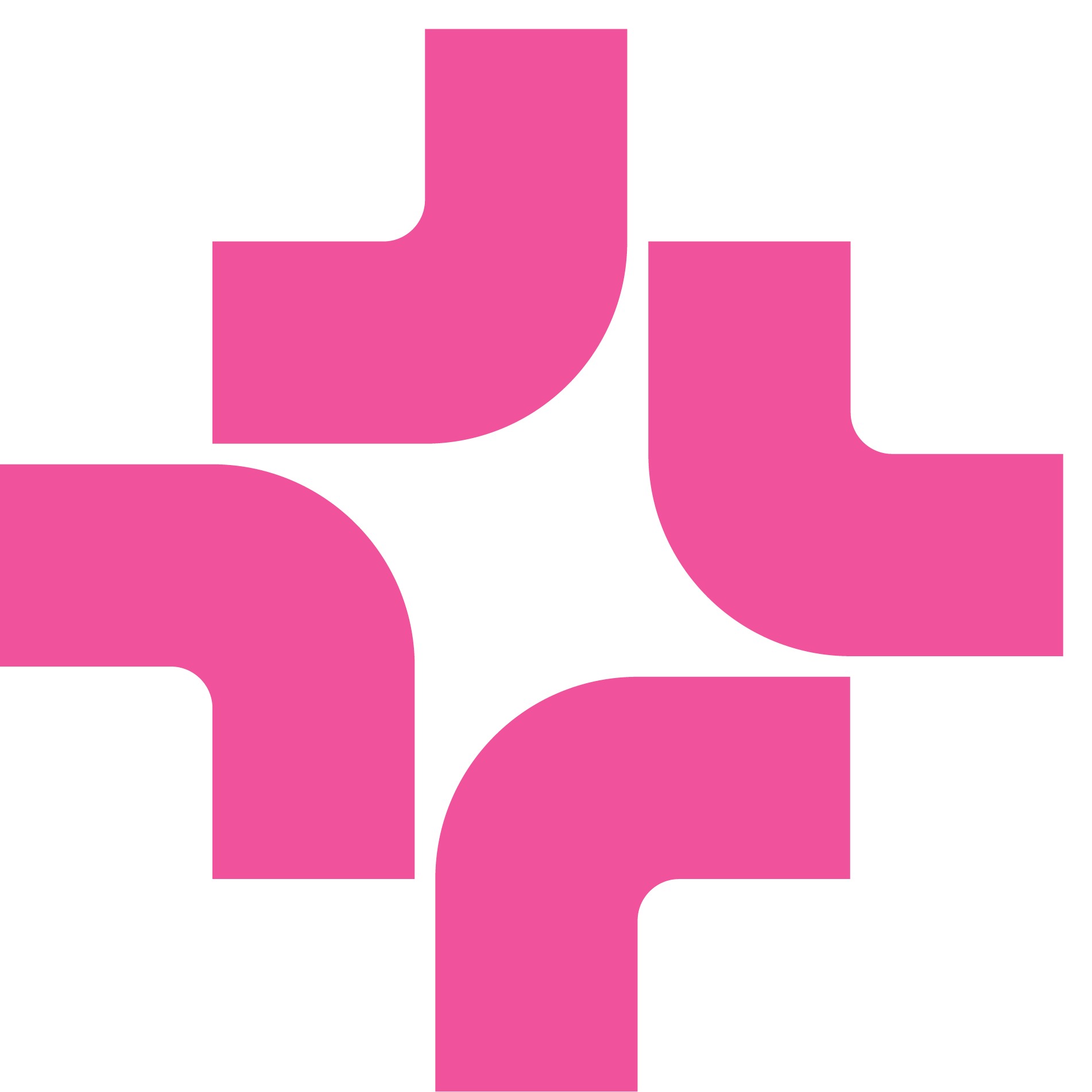}%
  \endgroup
  \hspace{1pt}%
}

\lstdefinestyle{mystyle}{
    basicstyle=\ttfamily\small,
    breaklines=true,
    frame=single,
    backgroundcolor=\color{gray!5},
    keywordstyle=\color{black},
    commentstyle=\color{gray},
    stringstyle=\color{black},
    showstringspaces=false,
    tabsize=2,
    rulecolor=\color{black},
    framesep=6pt, 
    columns=fullflexible,
}
\AtBeginDocument{
  \setlength{\abovedisplayskip}{4pt}
  \setlength{\belowdisplayskip}{4pt}
  \setlength{\abovedisplayshortskip}{2pt}
  \setlength{\belowdisplayshortskip}{2pt}
}

\title{Language Models as Higher-Order Planning Formalizers}

\author{Owen Jiang \drexel \enspace Cassie Huang \drexel \enspace Ashish Sabharwal \aitwo \enspace Li Zhang \drexel \\
  \drexel Drexel University \aitwo Allen Institute for AI \hspace{4pt} \\
  {{\tt {owenjiang669}@gmail.com}\ | {\tt {harry.zhang}@drexel.edu}}
}

\begin{document}


\setcounter{page}{1}

\maketitle

\begin{abstract}
Recent work provides overwhelming evidence that LLMs, even those trained to scale their reasoning trace, quickly deteriorate at planning as problems become more complex. \emph{LLM-as-Formalizers} aim to address this by employing LLMs as a bridge to translate natural language descriptions into structured planning representations such as PDDL, which are then fed to a programmatic solver. We observe that its success may be overstated because planning problem descriptions in standard benchmarks often have a one-to-one mapping to PDDL, which departs from real use cases. To address this, we introduce the notion of \emph{unraveling problems} where a natural yet succinct description translates into a very large PDDL representation.
Using unraveling variants of four standard planning domains, we demonstrate that LLM Formalizers also do not always scale. We tackle this challenge by introducing a new paradigm, \textit{LLM-as-Higher-Order-Formalizer}, where the LLM generates a high-level program that captures the recurrent logic within the description and in turn generates the larger PDDL representation. This decouples token output from the combinatorial explosion of the underlying formalization and search space, leading to improved performance for complex problems.\footnote{https://github.com/oj79/formalizer-scaling.git}
\end{abstract}

\section{Introduction}
Recent large language models (LLMs) have been advocated to be able to plan, or to be trained to plan. The most intuitive paradigm, \textit{LLM-as-Planner}, generates action sequences directly from domain and problem specifications in an end-to-end fashion. However, recent work provides overwhelming evidence that LLMs, even those trained to scale their reasoning trace (LRMs), collapse when solving planning problems beyond a certain complexity \cite{valmeekam2024llmscantplanlrms,shojaee2025illusionthinkingunderstandingstrengths,lin2025zebralogic,10.1007/978-3-032-11402-0_9, vyas2025are}. 
Additionally, they lack verifiability and interpretability.
An alternative paradigm, called \textit{LLM-as-Formalizer}, has been introduced as a remedy that employs the LLM as a bridge to translate natural language (NL) descriptions into structured formal representations like the Planning Domain Definition Language (PDDL) \cite{mcdermott1998pddl,huang-zhang-2025-limit}. 
The output is then fed to a programmatic planner to derive a plan. This provides built-in verifiability and is effective on standard benchmarks \cite{huang2025languagemodelplannerformalizer,kagitha2025unifyinginferencetimeplanninglanguage}.

\begin{figure}[t!]
    \centering
    \includegraphics[width=\columnwidth]{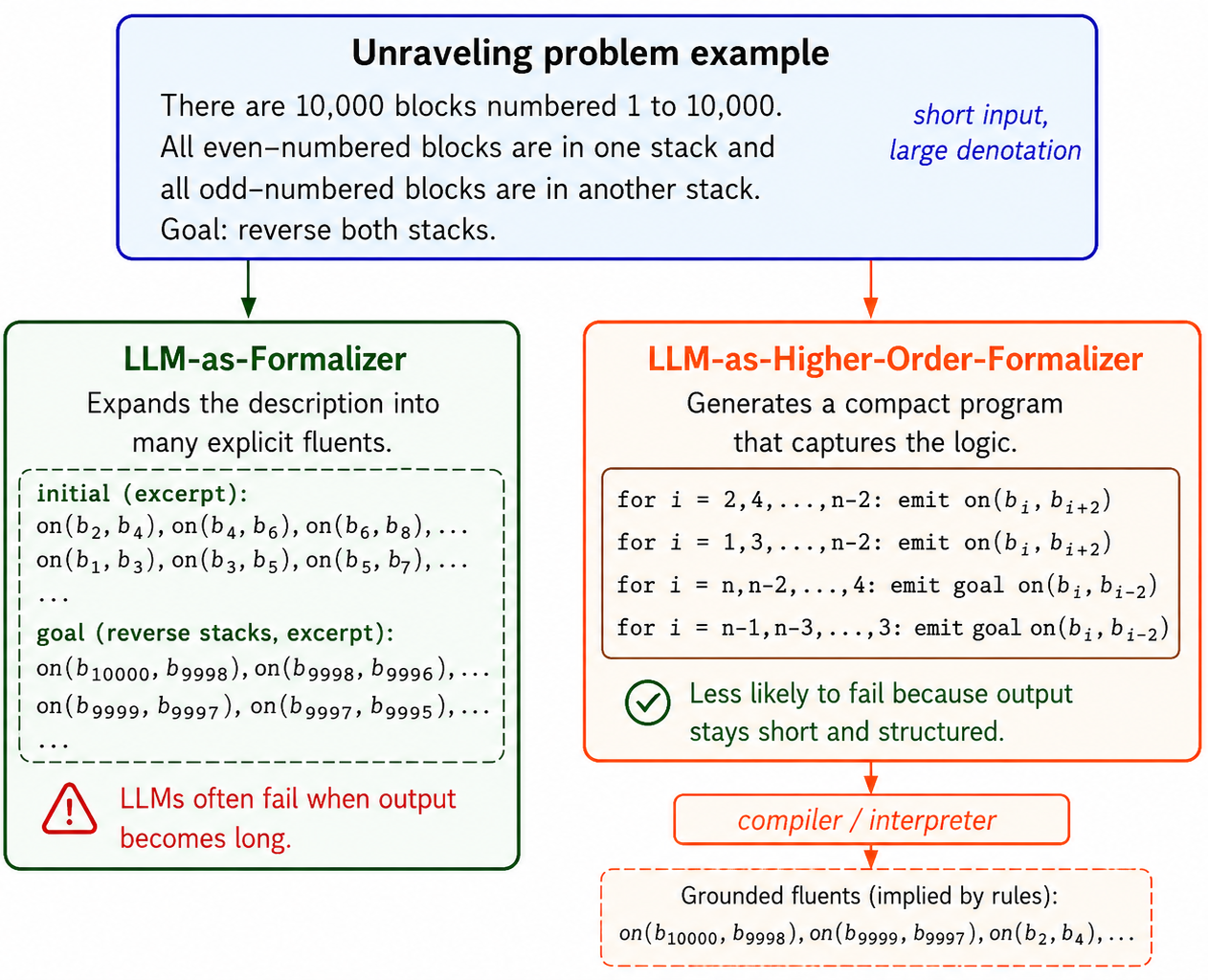}
    \caption{On an unraveling problem where the input description is short but the denoted planning instance is large, an LLM-as-Formalizer must explicitly enumerate many grounded fluents, while an LLM-as-Higher-Order-Formalizer generates a compact program that captures the same logic and can be expanded downstream.}
    \label{fig:unraveling}
\end{figure}

However, we observe that the success of the Formalizer paradigm may be overstated due to small problem sizes in classical planning domains. Therefore, \textbf{we significantly expand the problem sizes and already observe performance degradation in both Planner and Formalizer}. Then, we think that those expanded problems also reflect an unrealistic design of standard benchmarks, namely that the natural language problem description and the ground-truth PDDL code often have a one-to-one mapping. In industrial uses cases, succinct problem descriptions often describe large problems. For example, in warehouse storage, it is common to describe inventory as ``Zones 3306-8901..., shelves F-K..., or the first 10000 carts...''. Figure~\ref{fig:unraveling} also includes a classical planning domain example. In Figure~\ref{fig:unraveling}, the description ``There are 10,000 blocks, of which all even-numbered ones are in one stack.'' naturally translates to tens thousands of lines of PDDL. This is not reflected in any current benchmark, which is small in scale .

To remedy this, we first introduce the concept of unraveling problems where a single line of description corresponds to arbitrarily many lines in PDDL. We propose the \unravel benchmark based on four classical planning domains. \textbf{Our results reveal that Formalizer often degrades. Even our best frontier models sharply degrade more than 50\% and sometimes 100\%}.
How, then, can we design a LLM-based planning method that scale more reliably? To tackle this challenge, we introduce a new planning paradigm called LLM-as-Higher-Order-Formalizer. Here the LLM's job is to translate the natural language problem description into a \emph{high-level program} (e.g., in Python), which in turn generates the PDDL representation, which is then fed to a programmatic planner. Importantly, this high-level program is expected to roughly be in a one-to-one correspondence with the natural language description, capturing its recursive nature, and drastically shortening the token output of the LLM on compactly described problems. \textbf{We find that under this new paradigm, our best frontier models scale perfectly on \unravel, while weaker models achieve notably improved performance}.

The earlier Formalizer resembles code execution, requiring demanding state tracking~\cite{merrill-etal-2026-olmohybrid}. As LLMs become adept at coding, H-O Formalizer instead uses code generation to expand natural succinct descriptions into large-scale planning representations.

\section{LLM Planning and Scaling}
\label{sec:scaling-theory}
We start with theoretical discussions of the input, output, and semantic spaces an LLM navigates in different planning paradigms. We then present empirical results about how these paradigms fare when scaling problem size.

\subsection{In Theory}

Let \(D_n\) denote a natural-language description of a planning problem with \(n\) objects. 

\begin{definition}
Let
\[
   \mathcal{I}_n =  U(D_n)=\langle O_n, \mathcal{R}, F_n, A_n, I_n, G_n \rangle
\]
be the symbolic planning instance denoted by the natural-language
description \(D_n\), where:
\begin{itemize}
    \item \(O_n\) is a set of objects;
    \item \(\mathcal{R}\) is a finite set of predicate symbols;
    \item \(F_n\) is the set of ground fluents induced by \(O_n\) and
    \(\mathcal{R}\);
    \item \(A_n\) is the set of grounded actions;
    \item \(I_n \subseteq F_n\) is the initial state;
    \item \(G_n\) is the goal condition.
\end{itemize}
A state is a subset \(s \subseteq F_n\). A plan is a finite action
sequence $\pi = (a_1,a_2,\ldots,a_L), \qquad a_i \in A_n$
that transforms \(I_n\) into a state satisfying \(G_n\). 
\end{definition}

Therefore, the \textbf{task of planning} involves:
\[
    D_n
    \xrightarrow{\text{}}
    \mathcal{I}_n
    =
    \langle O_n,F_n,A_n,I_n,G_n\rangle
    \xrightarrow{\text{}}
    \pi.
\]


\paragraph{Spaces Induced by Planning Instances.}

A symbolic planning instance \(\mathcal{I}_n\) induces several distinct
spaces described next and not to be conflated:
\[
    \underbrace{F_n}_{\text{fluent space}},
    \underbrace{A_n}_{\text{action space}},
    \underbrace{\mathcal{S}_n }_{\text{state space}},
    \underbrace{\mathcal{T}_n}_{\text{search space}},
    \underbrace{A_n^L}_{\text{plan space}}.
\]

The fluent space \(F_n\) is a set of ground predicates forming the
symbolic vocabulary over which states are described. The state space
\(\mathcal{S}_n\) is the set of legal world configurations (where each world configuration is a state \(s \subseteq F_n\) is the subset of fluents that are true in that state). The action
space \(A_n\) contains grounded actions that map states to other states.
Actions are described by schemas, e.g., transport package 5 from location 10 to location 15. The search space
\(\mathcal{T}_n\) is the transition graph induced by states and actions
$\mathcal{T}_n = (\mathcal{S}_n, E_n)$,
where an edge $s \xrightarrow{a} s'$
exists if action \(a \in A_n\) is applicable in state
\(s \in \mathcal{S}_n\) and produces successor state
\(s' \in \mathcal{S}_n\). The plan space \(A_n^L\) contains
candidate action sequences of length \(L\). A valid plan is a path in
\(\mathcal{T}_n\) from the initial state \(I_n\) to a state satisfying the
goal \(G_n\).

For a given planning domain, if \(r\) is the fluent predicate arity
and \(\ell\) the action-schema arity,\footnote{For ease of exposition, we assume fixed arity for both.} then
\[
    |F_n| = \Theta(n^r), \ \ 
    |A_n| = \Theta(n^\ell).
\]
Thus the grounded \emph{fluent} and \emph{action vocabularies} are of size polynomial in \(n\). In contrast, the \emph{state space} is exponential in
the fluent vocabulary,
\[
    |S_n| = 2^{|F_n|} = 2^{\Theta(n^r)},
\]
and the \emph{plan space} of length \(L\) is exponential in \(L\)
\[
    |A_n^L| = |A_n|^L = \Theta(n^{\ell \cdot L}).
\]
The \emph{search space} is the transition graph constrained by action applicability and goal satisfaction, which we approximate as 
\[
    |\mathcal{T}_n| \approx \Theta(|\mathcal{S}_n|\,|A_n|).
\]

\paragraph{LLM-as-Planner.}

\begin{figure}[t!]
    \centering
    \includegraphics[width=\columnwidth]{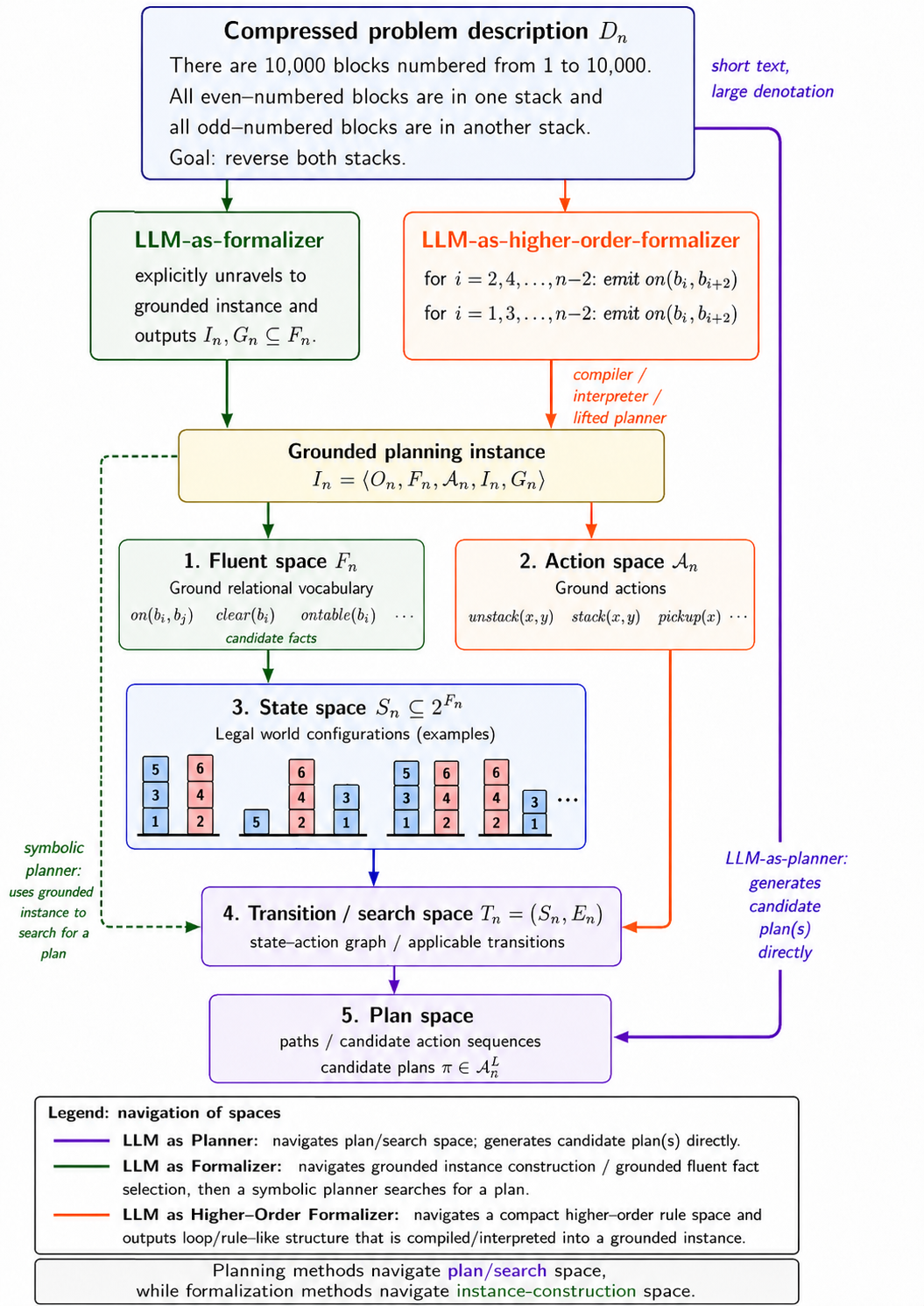}
    \caption{An illustration of the nature of combinatorial explosion in the BlocksWorld domain. Under different paradigms, LLMs generate tokens tantamount to the complexity of different spaces.}
    \label{fig:explosion}
\end{figure}
An LLM may directly map the natural-language description to a
plan by generating just the plan $D_n \mapsto \widehat{\pi}$.
The generated output \(\widehat{\pi}\) is a token sequence, but a direct planner must implicitly perform two
tasks: translate \(D_n\) to the planning instance $\mathcal{I}_n$ and navigate the \emph{search space \(\mathcal{T}_n\)}, which is often infeasible for an inference pass that runs in time $O(n^2)$ for a transformer LLM.

\paragraph{LLM-as-Planner with Inference-Time Scaling.}

Inference-time scaling gives the LLM a larger budget to explore candidate
outputs $\widehat{\pi}_1,\widehat{\pi}_2,\ldots,\widehat{\pi}_B$.
At the token level, this expands the explored generation space. At the semantic level, these outputs correspond to candidate paths within the \textit{same} \emph{underlying space} \(\mathcal{T}_n\). Therefore, inference-time scaling changes the amount of search performed,
not the nature of the search problem. 

\paragraph{LLM-as-Formalizer.}

The Formalizer does not directly navigate the search space. Instead,
given a fixed domain, it navigates the \emph{space of ground fluents} \(F_n\), which scales polynomially in \(n\).
The combinatorial plan search is
delegated to a symbolic planner. See Figure~\ref{fig:explosion}. 

\subsection{In Practice}
\label{sec:scaling_practice}

We follow \citet{huang-zhang-2025-limit} for the concrete task formulation: given a domain specification (describes actions and their pre-conditions and effects) both in NL and PDDL and a problem description (describes initial and goal states of entities), a system outputs a plan (a sequence of actions) which is validated via a ground-truth simulation. Planner does so in an end-to-end manner, while Formalizer does so by generating a PDDL problem file which a programmatic planner takes in and produces a plan. We assume the domain file as given, following works like \citet{zuo-etal-2025-planetarium} as we only consider the scaling of problem complexity, where the domain remains constant.
We normally use the \texttt{dual-bfws-ffparser} planner implemented by \citet{muise-icaps16demo-pd} as the programmatic planner and VAL \cite{1374201} as the validator. However, since the programmatic planner times-out as problems grow in complexity, we write our own parsers to compare the LLM-generated PDDL problem files with the ground-truth problem files for differences. Since BlocksWorld guaranties solvability, if the parser shows that the LLM-generated problem files perfectly match the ground-truths, we count that as providing an accurate plan. We report the plan \textit{accuracy}, the percentage that can be executed to achieve the goal state.

\paragraph{Benchmark.}
We focus on the arguably most used planning domains in the community, BlocksWorld \cite{ipc} based on object manipulation to rearrange stacks of blocks on a table. We extend an existing dataset from \citet{huang-zhang-2025-limit} with ``moderately templated'' descriptions but create problems with much higher complexity than existing work (see Appendix~\ref{app:extension}). Examples of the NL domain and problem descriptions, the Planner and Formalizer prompts, and the PDDL domain and problem files are shown in the Appendix starting Listing \ref{lst:blocksworld_df}. The resulting BlocksWorld-XXL dataset includes 200 problems ranging from 5 to 100 blocks. We use the size of the object space as a proxy of problem complexity. 


\paragraph{Models.}
We consider Gemini 3 Flash (\texttt{G3F}) as a representative, state-of-the-art, closed-source LRM, DeepSeek-V4-Flash (\texttt{DS-V4}) \footnote{This has been re-routed to DeepSeek-V4.1-Flash since September 10, 2026} \cite{deepseekai2026deepseekv4} as an open-source LRM , and Qwen2.5-Coder-32B-Instruct (\texttt{Q25}) \cite{hui2024qwen2} as an open-source LLM. Open models are run using KANI \cite{zhu-etal-2023-kani} with default temperature on 1 H100 GPU. 

\paragraph{Scaling Performance.}
\label{sec:scaling-results}
Figure~\ref{fig:merged_scaling_curves_all}a shows the scaling performance of 3 models as both Planners and Formalizers on BlocksWorld-XXL. We conclusively corroborate previous work that Planners do not scale with problem complexity. All models including the LRMs trained to scale reasoning token degrade to 20\% accuracy or worse as Planners at merely 30 blocks, while both LRMs display a ``spark'' of planning ability at only 5 blocks. In contrast, Formalizers scale much better with problem complexity. The performance of \texttt{G3F} remains 100\% up to 100 blocks, while that of the non-reasoning \texttt{Q25} remains above 70\% until $80$ blocks but sharply degrades thereafter. Error analysis on \texttt{Q25}'s outputs shows 14\% have missing initial conditions, 64\% have extra initial conditions, 57\% have missing goals, and 21\% have extra goals. 


To show weaker models' inability to formalize long statements, we propose a \textbf{divide-and-conquer} (D\&C) technique.
We prompt an LLM to first generate the problem file headers with one call, including problem name, domain name, and objects. Next, the input problem description is segmented into sentences. Provided one sentence at a time plus the domain specification to avoid context overload, the LLM generates only one line of PDDL code. Since each generated line is grounded in the domain file and thus independent of each other, the lines are later consolidated. Figure~\ref{fig:merged_scaling_curves_all}a shows that D\&C addresses the context overloading problem for weaker models, increasing \texttt{Q25} performance as Formalizer from 30\% to 100\% at 100 blocks. 


\section{Uraveling Problems}
\label{sec: unraveling problems}
In all existing planning benchmarks juxtaposing NL and PDDL, the problem descriptions $D_n$ and problem files $(O_n,F_n,G_n)$ are aligned as a one-to-one, line-to-line mapping, so that formalizing them may be oversimplifying real-life problems. As an extreme example in theorem proving, the Kepler conjecture is a single statement that took 11 years to formalize \cite{hales2015formalproofkeplerconjecture}.

\subsection{Definition}
We call a planning problem an \textbf{unraveling problem} when the natural
language description \(D_n\) is succinct, but the denoted planning instance
\[
    U(D_n) = \mathcal{I}_n
    =
    \langle O_n,F_n,A_n,I_n,G_n\rangle
\]
is significantly larger ($|D_n| \ll |\mathcal{I}_n|$). This creates a distinct scaling challenge for a Formalizer that must still explicitly emit polynomially many
facts from the compact description.

To address this compression gap, we introduce the paradigm of
\textbf{LLM-as-Higher-Order-Formalizer}. Instead of directly outputting the
fully grounded instance
\(\mathcal{I}_n\),
the model outputs a
higher-order generator program first:
\[
    D_n \mapsto R_n \mapsto \mathcal{I}_n,
\]
where
\(R_n\)
compactly describes (e.g., in Python) how to emit the objects and
fluents of the instance.
A downstream compiler, interpreter, or lifted planner then expands this
higher-order representation into a grounded planning instance.
While a regular Formalizer navigates the ground fluent space $F_n$, 
an H-O Formalizer instead navigates a compact program
space; its generated text can remain
small even when the denoted instance is large
($|R_n| \ll |F_n|$).
This shifts Formalizer's burden from enumerating large grounded fluent sets to producing a compact generation rule of them.

\subsection{Unraveling Datasets}
We create the first benchmark of unraveling problems coined \unravel based on 4 classical planning domains: BlocksWorld \cite{ipc}, Elevators \cite{ipc-2008}, Transport \cite{ipc-2008}, and ChildSnack \cite{ipc-2014}. Our data creation process can be extended to any similar planning domain.

\paragraph{Unraveling Problems Layout.}
In unraveling problems, the most important distinction is that we no longer describe problems line-by-line, but with compact language that portrays the overall properties and relations of the objects in the domains, effectively compressing the problem descriptions (Figure~\ref{fig:unraveling}).
As an example from ChildSnack, an enumerative description like \emph{`There are  child 1, child 2, child 3, child 4, child 5, tray 1, tray 2, tray 3, bread portion 1, bread portion 2, bread portion 3, $\ldots$'} turns into a compressed, higher-level description \emph{`There are $n$ children numbered 1 to $5$, bread portions numbered 1 to 3, $\ldots$'.}  Therefore, the length of the the natural language description remains roughly constant as the problem complexity scales up. 



\paragraph{Creating Unraveling Problems.}
We create problem descriptions by picturing a realistic scenario with an easily describable pattern for each domain. For example, in ChildSnack, we use a controlled serving scenario in which children are partitioned into groups waiting at fixed places, with one tray serving each group. Unlike standard ChildSnack instances, where the distribution of children, ingredients, trays, and dietary requirements can vary more freely, our instances assign each child a corresponding bread portion, content portion, and sandwich placeholder, and use a deterministic regular/gluten-free dietary pattern with matching gluten-free ingredients. The goal is to prepare and serve one appropriate sandwich to every child. We refer to this scenario as `single sandwich line' because each child appears in exactly one serving line, although an instance may contain multiple tray-specific lines. 


\paragraph{Choosing Main-Varying versus Fixed Parameters.}
All domains except BlocksWorld have multiple variable parameters, such as the number of children, trays, and content portions in ChildSnack. Theoretically, we could randomly vary all parameters to create distinct problems. However, that makes it impossible to systematically study how LLMs' performance changes for more, or less complex problems. Therefore, we choose a main parameter in each domain to vary, where the values that the main parameter takes on is $n$. We fix other parameters at a medium-level difficulty based on the original competitions. For example, in ChildSnack, we choose the number of children as the main parameter and fix the numbers of trays to 3 versus a maximum of 4 in the original competition. The main and fixed parameters for Transport and Elevators are shown in Appendix \ref{app:main_fixed_parameters}.  

\begin{figure}[t!]
    \centering
    \includegraphics[width=\columnwidth]{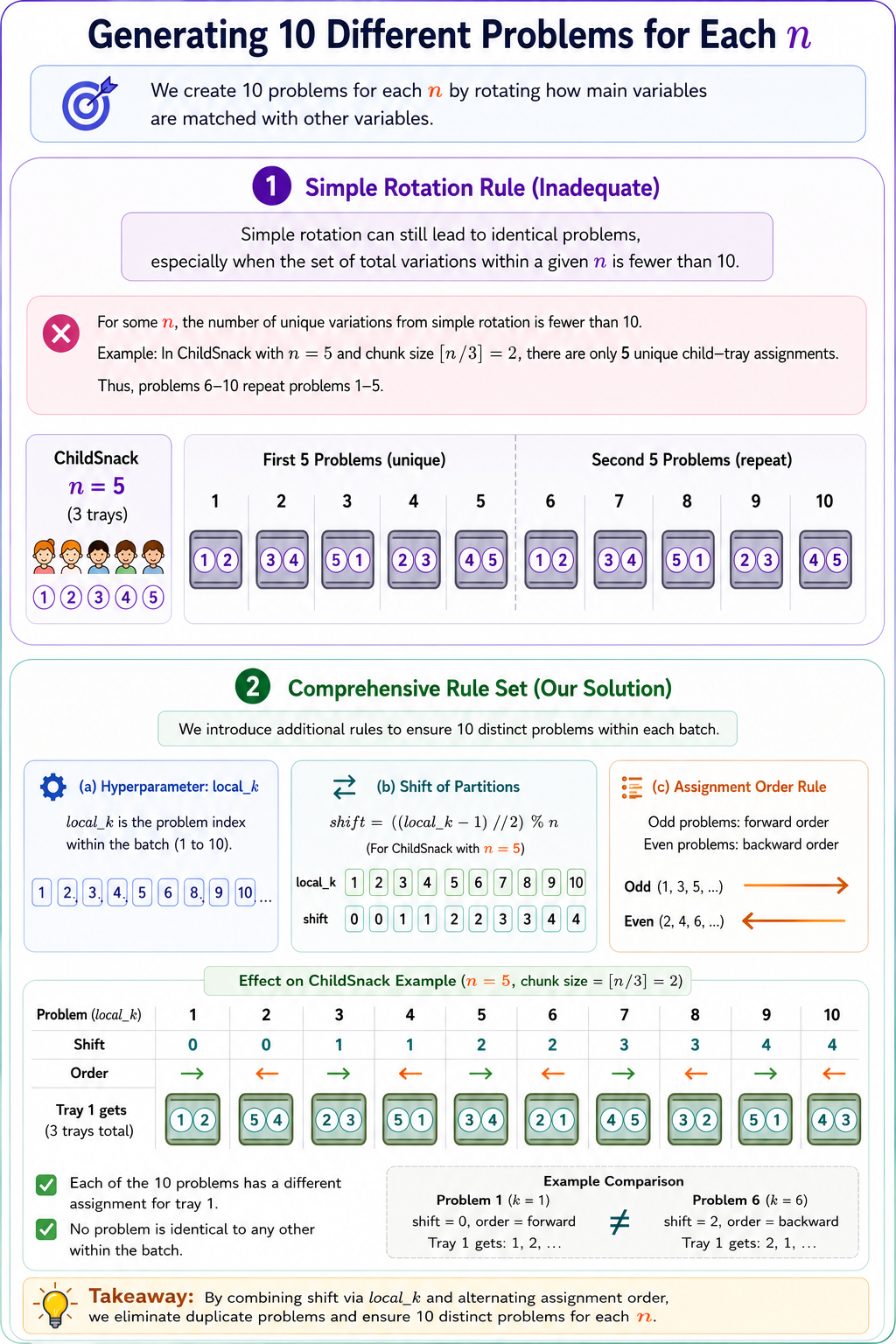}
    \caption{Using a combination of rules to create distinct unraveling problems in ChildSnack.}
    \label{fig:distinct_problems}
\end{figure}

\paragraph{Ensuring Distinct Within-Batch Problems.}
We create 10 problems for each $n$ per domain. We call each 10-problem group a batch. To create distinct problems within each batch, we rotate the way the main-varying parameters are matched with the fixed parameters. For example, we rotate the trays that the children are assigned in ChildSnack. However, a simple partition, such as grouping children using a fixed division rule like $ceil(n / 3)$ can still lead to identical problems within a batch. For example, in the $n=5$ batch of ChildSnack, there are only five unique child-tray assignments if we only use a primitive partition formula $children \ group = ceil(n / 3)$. In a problem with five children, the second five problems will repeat the partitions of the first five. To solve this, we first introduce a "hyperparameter" $local\_k$ that is equivalent to the problem index within a batch. This hyperparameter helps determine the partition shift for each problem within the batch using $shift = ((local\_k - 1) // 2) \ \% \ n$. Furthermore, we add an assignment order rule where odd problems use forward child order and even problems use backward child order. These designs help eliminate identical problems within a given batch. Figure \ref{fig:distinct_problems} provides an example of combining rules to create distinct problems. 


\paragraph{Dataset Statistics.}
The \unravel dataset contains 270 problem descriptions and PDDL files for four domains, including problem descriptions and PDDL files ranging from 5 to 100 blocks for BlocksWorld, from 10 to 100 floors for Elevators, from 10 to 100 locations for Transport, and from 5 to 100 children for ChildSnack. The dataset enables us to draw convincing conclusions, while it can be expanded without costs by adjusting the numerical expressions like $n$. Note that as we scale up problem complexity (both entity space and fluent space), the input size of the unraveling problems remains constant. 

\subsection{Implementing H-O Formalizers}
We conduct comprehensive experiments for all domains and pipelines. In H-O Formalizer, we provide LLMs with NL domain and problem descriptions, H-O Formalizer prompts, and ground-truth PDDL domain files and ask the LLMs to generate Python programs that produce PDDL problem files upon execution. Examples of the domain and problem descriptions, H-O Formalizer prompts, PDDL domain and problem files are shown in Appendix starting from Listing \ref{lst:blocksworld_higher_order_formalizer_prompt}. Like in Formalizer, we also use our own parsers to perform differences checking since all four domains guaranties solvability. In H-O Formalizer, inspired by the fact that the LLMs naturally need to write code for repeating patterns such as loops or appending object types to formalize the problems, we use two-stage prompts where we ask the LLMs to reflect on the repeating patterns that they just wrote for the first prompt before finalizing the Python programs, especially paying attention to possible errors in the way problem descriptions translate to repeating patterns. For fairness of comparison, we do the same with Formalizer where the review focuses on PDDL errors instead of repeating patterns. 


\begin{figure*}[p]
\centering
\setcounter{subfigure}{0}

\pgfplotsset{
    mergedaxis/.style={
        width=0.95\linewidth,
        height=3.9cm,
        ymin=0, ymax=100,
        ytick={0,20,40,60,80,100},
        yticklabel={\pgfmathprintnumber{\tick}\%},
        tick label style={font=\tiny},
        xticklabel style={
            font=\tiny,
            rotate=60,
            anchor=east
        },
        title style={font=\scriptsize\bfseries, yshift=-1ex},
        grid=major,
        grid style={dashed, gray!30}
    }
}

\refstepcounter{subfigure}\label{fig:blocks_xxl}
\textbf{\footnotesize(\thesubfigure) BlocksWorld-XXL}\par\vspace{1mm}

\begin{minipage}{0.32\textwidth}
\begin{tikzpicture}
\begin{axis}[mergedaxis,title={As Planner},xmin=0,xmax=100,xtick={5,30,50,75,100}]
\addplot[color=blue, mark=square*, mark size=1.5pt, thick] coordinates {(5,70) (30,20) (50,10) (75,10) (100,0)};
\addplot[color=red!60!black, mark=*, mark size=1.5pt, thick] coordinates {(5,80) (30,30) (50,30) (75,20) (100,0)};
\addplot[color=green!60!black, mark=*, mark size=1.5pt, thick] coordinates {(5,20) (30,0) (50,0) (75,10) (100,0)};
\end{axis}
\end{tikzpicture}
\end{minipage}%
\hfill
\begin{minipage}{0.32\textwidth}
\begin{tikzpicture}
\begin{axis}[mergedaxis,title={As Formalizer},xmin=0,xmax=100,xtick={5,30,50,75,100},yticklabels={}]
\addplot[color=blue, mark=square*, mark size=1.5pt, thick] coordinates {(5,100) (30,100) (50,100) (75,100) (100,100)};
\addplot[color=red!60!black, mark=*, mark size=1.5pt, thick] coordinates {(5,90) (30,90) (50,80) (75,80) (100,80)};
\addplot[color=green!60!black, mark=*, mark size=1.5pt, thick] coordinates {(5,100) (30,80) (50,80) (75,70) (100,30)};
\end{axis}
\end{tikzpicture}
\end{minipage}%
\hfill
\begin{minipage}{0.32\textwidth}
\begin{tikzpicture}
\begin{axis}[mergedaxis,title={As D\&C Formalizer},xmin=0,xmax=100,xtick={5,30,50,75,100},yticklabels={}]
\addplot[color=green!60!black, mark=*, mark size=1.5pt, thick] coordinates {(5,100) (30,100) (50,100) (75,100) (100,100)};
\end{axis}
\end{tikzpicture}
\end{minipage}

\vspace{2mm}

\hrule

\vspace{2mm}

\refstepcounter{subfigure}\label{fig:blocksworld_unravel}
\textbf{\footnotesize(\thesubfigure) BlocksWorld-Unravel}\par\vspace{1mm}

\begin{minipage}{0.32\textwidth}
\begin{tikzpicture}
\begin{axis}[mergedaxis,title={As Planner},xmin=0,xmax=100,xtick={0,20,40,60,80,100}]
\addplot[color=blue, mark=square*, mark size=1.5pt, thick] coordinates {(5,0) (30,0) (50,0) (75,10) (100,0)};
\addplot[color=green!60!black, mark=*, mark size=1.5pt, thick] coordinates {(5,0) (30,0) (50,0) (75,0) (100,0)};
\addplot[color=red!60!black, mark=*, mark size=1.5pt, thick] coordinates {(5,20) (30,0) (50,0) (75,10) (100,0)};
\end{axis}
\end{tikzpicture}
\end{minipage}%
\hfill
\begin{minipage}{0.32\textwidth}
\begin{tikzpicture}
\begin{axis}[mergedaxis,title={As Formalizer},xmin=0,xmax=100,xtick={0,20,40,60,80,100},yticklabels={}]
\addplot[color=blue, mark=square*, mark size=1.5pt, thick] coordinates {(5,100) (30,100) (50,100) (75,100) (100,100)};
\addplot[color=green!60!black, mark=*, mark size=1.5pt, thick] coordinates {(5,60) (30,0) (50,0) (75,0) (100,0)};
\addplot[color=red!60!black, mark=*, mark size=1.5pt, thick] coordinates {(5,100) (30,100) (50,100) (75,100) (100,100)};
\addplot[color=black!60!black, mark=*, mark size=1.5pt, thick] coordinates {(5,60) (30,10) (50,0) (75,0) (100,0)};
\end{axis}
\end{tikzpicture}
\end{minipage}%
\hfill
\begin{minipage}{0.32\textwidth}
\begin{tikzpicture}
\begin{axis}[mergedaxis,title={As H-O Formalizer},xmin=0,xmax=100,xtick={0,20,40,60,80,100},yticklabels={}]
\addplot[color=blue, mark=square*, mark size=1.5pt, thick] coordinates {(5,100) (30,100) (50,100) (75,100) (100,100)};
\addplot[color=green!60!black, mark=*, mark size=1.5pt, thick] coordinates {(5,90) (30,70) (50,20) (75,10) (100,10)};
\addplot[color=brown!60!black, mark=*, mark size=1.5pt, thick] coordinates {(5,90) (30,60) (50,30) (75,30) (100,20)};
\addplot[color=red!60!black, mark=*, mark size=1.5pt, thick] coordinates {(5,100) (30,100) (50,100) (75,100) (100,100)};
\end{axis}
\end{tikzpicture}
\end{minipage}

\vspace{2mm}

\refstepcounter{subfigure}\label{fig:elevators_unravel}
\textbf{\footnotesize(\thesubfigure) Elevators-Unravel}\par\vspace{1mm}

\begin{minipage}{0.32\textwidth}
\begin{tikzpicture}
\begin{axis}[mergedaxis,title={As Planner},xmin=5,xmax=100,xtick={0,10,15,20,25,30,50,100}]
\addplot[color=blue, mark=square*, mark size=1.5pt, thick] coordinates {(10,0) (15,0) (20,10) (25,0) (30,0) (50,0) (100,0)};
\addplot[color=green!60!black, mark=*, mark size=1.5pt, thick] coordinates {(10,0) (15,0) (20,0) (25,0) (30,0) (50,0) (100,0)};
\addplot[color=red!60!black, mark=*, mark size=1.5pt, thick] coordinates {(10,0) (15,0) (20,0) (25,0) (30,10) (50,0) (100,0)};
\end{axis}
\end{tikzpicture}
\end{minipage}%
\hfill
\begin{minipage}{0.32\textwidth}
\begin{tikzpicture}
\begin{axis}[mergedaxis,title={As Formalizer},xmin=5,xmax=100,xtick={0,10,15,20,25,30,50,100},yticklabels={}]
\addplot[color=blue, mark=square*, mark size=1.5pt, thick] coordinates {(10,100) (15,100) (20,100) (25,100) (30,100) (50,70) (100,0)};
\addplot[color=green!60!black, mark=*, mark size=1.5pt, thick] coordinates {(10,0) (15,0) (20,0) (25,0) (30,0) (50,0) (100,0)};
\addplot[color=black!60!black, mark=*, mark size=1.5pt, thick] coordinates {(10,0) (15,0) (20,0) (25,0) (30,0) (50,0) (100,0)};
\addplot[color=red!60!black, mark=*, mark size=1.5pt, thick] coordinates {(10,100) (15,100) (20,100) (25,100) (30,100) (50,100) (100,50)};
\end{axis}
\end{tikzpicture}
\end{minipage}%
\hfill
\begin{minipage}{0.32\textwidth}
\begin{tikzpicture}
\begin{axis}[mergedaxis,title={As H-O Formalizer},xmin=5,xmax=100,xtick={0,10,15,20,25,30,50,100},yticklabels={}]
\addplot[color=blue, mark=square*, mark size=1.5pt, thick] coordinates {(10,100) (15,100) (20,100) (25,100) (30,100) (50,100) (100,100)};
\addplot[color=green!60!black, mark=*, mark size=1.5pt, thick] coordinates {(10,0) (15,0) (20,0) (25,0) (30, 20) (50, 10) (100,10)};
\addplot[color=brown!60!black, mark=*, mark size=1.5pt, thick] coordinates {(10,0) (15,20) (20,20) (25,10) (30,20) (50,10) (100,10)};
\addplot[color=red!60!black, mark=*, mark size=1.5pt, thick] coordinates {(10,100) (15,100) (20,100) (25,100) (30,100) (50,100) (100,100)};
\end{axis}
\end{tikzpicture}
\end{minipage}

\vspace{2mm}

\refstepcounter{subfigure}\label{fig:transport_unravel}
\textbf{\footnotesize(\thesubfigure) Transport-Unravel}\par\vspace{1mm}

\begin{minipage}{0.32\textwidth}
\begin{tikzpicture}
\begin{axis}[mergedaxis,title={As Planner},xmin=10,xmax=100,xtick={10,15,20,25,30,50,100}]
\addplot[color=blue, mark=square*, mark size=1.5pt, thick] coordinates {(10,0) (15,10) (20,0) (25,10) (30,0) (50,0) (100,0)};
\addplot[color=green!60!black, mark=*, mark size=1.5pt, thick] coordinates {(10,0) (15,0) (20,0) (25,0) (30,0) (50,0) (100,0)};
\addplot[color=red!60!black, mark=*, mark size=1.5pt, thick] coordinates {(10,0) (15,0) (20,0) (25,0) (30,0) (50,0) (100,0)};
\end{axis}
\end{tikzpicture}
\end{minipage}%
\hfill
\begin{minipage}{0.32\textwidth}
\begin{tikzpicture}
\begin{axis}[mergedaxis,title={As Formalizer},xmin=10,xmax=100,xtick={10,15,20,25,30,50,100},yticklabels={}]
\addplot[color=blue, mark=square*, mark size=1.5pt, thick] coordinates {(10,100) (15,100) (20,100) (25,100) (30,100) (50,40) (100,0)};
\addplot[color=green!60!black, mark=*, mark size=1.5pt, thick] coordinates {(10,90) (15,90) (20,30) (25,0) (30,0) (50,0) (100,0)};
\addplot[color=black!60!black, mark=*, mark size=1.5pt, thick] coordinates {(10,100) (15,100) (20,80) (25,30) (30,0) (50,0) (100,0)};
\addplot[color=red!60!black, mark=*, mark size=1.5pt, thick] coordinates {(10,100) (15,100) (20,100) (25,100) (30,90) (50,90) (100,40)};
\end{axis}
\end{tikzpicture}
\end{minipage}%
\hfill
\begin{minipage}{0.32\textwidth}
\begin{tikzpicture}
\begin{axis}[mergedaxis,title={As H-O Formalizer},xmin=10,xmax=100,xtick={10,15,20,25,30,50,100},yticklabels={}]
\addplot[color=blue, mark=square*, mark size=1.5pt, thick] coordinates {(10,100) (15,100) (20,100) (25,100) (30,100) (50,100) (100,100)};
\addplot[color=green!60!black, mark=*, mark size=1.5pt, thick] coordinates {(10,90) (15,90) (20,70) (25,60) (30,50) (50,50) (100,50)};
\addplot[color=brown!60!black, mark=*, mark size=1.5pt, thick] coordinates {(10,90) (15,70) (20,80) (25,90) (30,80) (50,70) (100,70)};
\addplot[color=red!60!black, mark=*, mark size=1.5pt, thick] coordinates {(10,100) (15,100) (20,100) (25,100) (30,100) (50,100) (100,100)};
\end{axis}
\end{tikzpicture}
\end{minipage}

\vspace{2mm}

\refstepcounter{subfigure}\label{fig:childsnack_unravel}
\textbf{\footnotesize(\thesubfigure) ChildSnack-Unravel}\par\vspace{1mm}

\begin{minipage}{0.32\textwidth}
\begin{tikzpicture}
\begin{axis}[mergedaxis,title={As Planner},xmin=5,xmax=100,xtick={5,10,15,20,25,30,50,100}]
\addplot[color=blue, mark=square*, mark size=1.5pt, thick] coordinates {(5,0) (10,10) (15,10) (20,30) (25,0) (30,0) (50,0) (100,0)};
\addplot[color=green!60!black, mark=*, mark size=1.5pt, thick] coordinates {(5,0) (10,0) (15,0) (20,0) (25,0) (30,0) (50,0) (100,0)};
\addplot[color=red!60!black, mark=*, mark size=1.5pt, thick] coordinates {(5,0) (10,10) (15,10) (20,10) (25,0) (30,10) (50,10) (100,0)};
\end{axis}
\end{tikzpicture}
\end{minipage}%
\hfill
\begin{minipage}{0.32\textwidth}
\begin{tikzpicture}
\begin{axis}[mergedaxis,title={As Formalizer},xmin=5,xmax=100,xtick={5,10,15,20,25,30,50,100},yticklabels={}]
\addplot[color=blue, mark=square*, mark size=1.5pt, thick] coordinates {(5,100) (10,100) (15,100) (20,100) (25,100) (30,100) (50,100) (100,90)};
\addplot[color=green!60!black, mark=*, mark size=1.5pt, thick] coordinates {(5,0) (10,40) (15,80) (20,70) (25,30) (30,50) (50,20) (100,10)};
\addplot[color=black!60!black, mark=*, mark size=1.5pt, thick] coordinates {(5,0) (10,70) (15,90) (20,70) (25,70) (30,70) (50,40) (100,30)};
\addplot[color=red!60!black, mark=*, mark size=1.5pt, thick] coordinates {(5,100) (10,100) (15,100) (20,90) (25,100) (30,100) (50,80) (100,100)};
\end{axis}
\end{tikzpicture}
\end{minipage}%
\hfill
\begin{minipage}{0.32\textwidth}
\begin{tikzpicture}
\begin{axis}[mergedaxis,title={As H-O Formalizer},xmin=5,xmax=100,xtick={5,10,15,20,25,30,50,100},yticklabels={}]
\addplot[color=blue, mark=square*, mark size=1.5pt, thick] coordinates {(5,100) (10,100) (15,100) (20,100) (25,100) (30,100) (50,100) (100,100)};
\addplot[color=green!60!black, mark=*, mark size=1.5pt, thick] coordinates {(5,20) (10,20) (15,0) (20,30) (25,20) (30,20) (50,40) (100,0)};
\addplot[color=brown!60!black, mark=*, mark size=1.5pt, thick] coordinates {(5,50) (10,40) (15,50) (20,40) (25,50) (30,50) (50,80) (100,50)};
\addplot[color=red!60!black, mark=*, mark size=1.5pt, thick] coordinates {(5,100) (10,100) (15,100) (20,100) (25,100) (30,100) (50,100) (100,100)};
\end{axis}
\end{tikzpicture}
\end{minipage}

\vspace{1mm}

\begin{tikzpicture}
\begin{axis}[
    hide axis,
    xmin=0, xmax=1,
    ymin=0, ymax=1,
    legend columns=-1,
    legend style={
        draw=none,
        font=\footnotesize,
        /tikz/every even column/.append style={column sep=0.35cm}
    }
]
\addlegendimage{color=blue, mark=square*, mark size=1.5pt, thick}
\addlegendentry{Gemini 3 Flash}

\addlegendimage{color=green!60!black, mark=*, mark size=1.5pt, thick}
\addlegendentry{Qwen2.5-32B}

\addlegendimage{color=brown!60!black, mark=*, mark size=1.5pt, thick}
\addlegendentry{+Pattern Review}

\addlegendimage{color=black!60!black, mark=*, mark size=1.5pt, thick}
\addlegendentry{+Plain Review}

\addlegendimage{color=red!60!black, mark=*, mark size=1.5pt, thick}
\addlegendentry{DeepSeek-V4-Flash}
\end{axis}
\end{tikzpicture}

\caption{a. Performance of state-of-the-art LLM Planners and Formalizers, on BlocksWorld with increasing problem complexity quantified by entity space size. For the divide-and-conquer (D\&C) technique, we omit Gemini 3 Flash and DeepSeek-V4-Flash due to their already high accuracy as regular formalizers; 
b--e. Performance of state-of-the-art LLMs Planners, Formalizers, and H-O Formalizers on \unravel across four classical planning domains with increasing problem complexity quantified by input size.
}
\label{fig:merged_scaling_curves_all}

\end{figure*}

\section{Results}
\begin{figure}[t!]
    \centering
    \includegraphics[width=0.9\columnwidth]{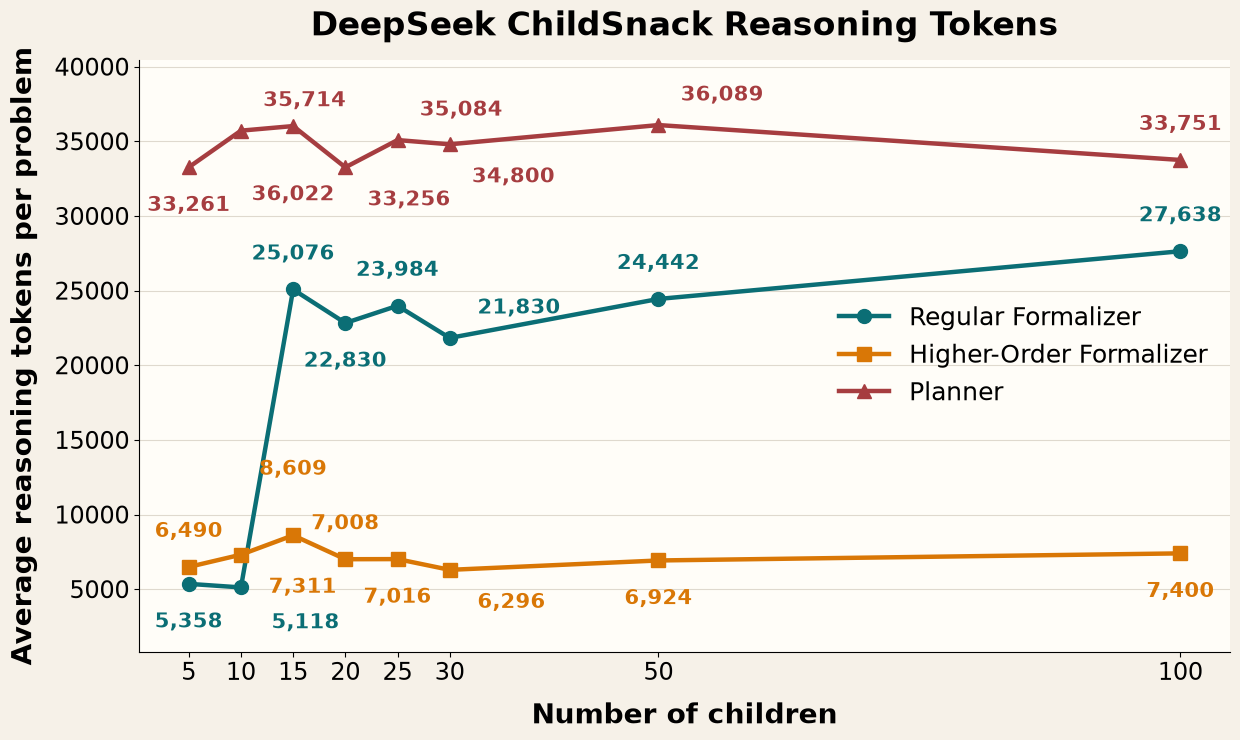}
    \caption{H-O Formalizer significantly reduces reasoning token consumption in ChildSnack.}
    \label{fig:childsnack_reasoning_tokens}
    \vspace{-2ex}
\end{figure}

\begin{figure}[t!]
    \centering
    \includegraphics[width=0.9\columnwidth]{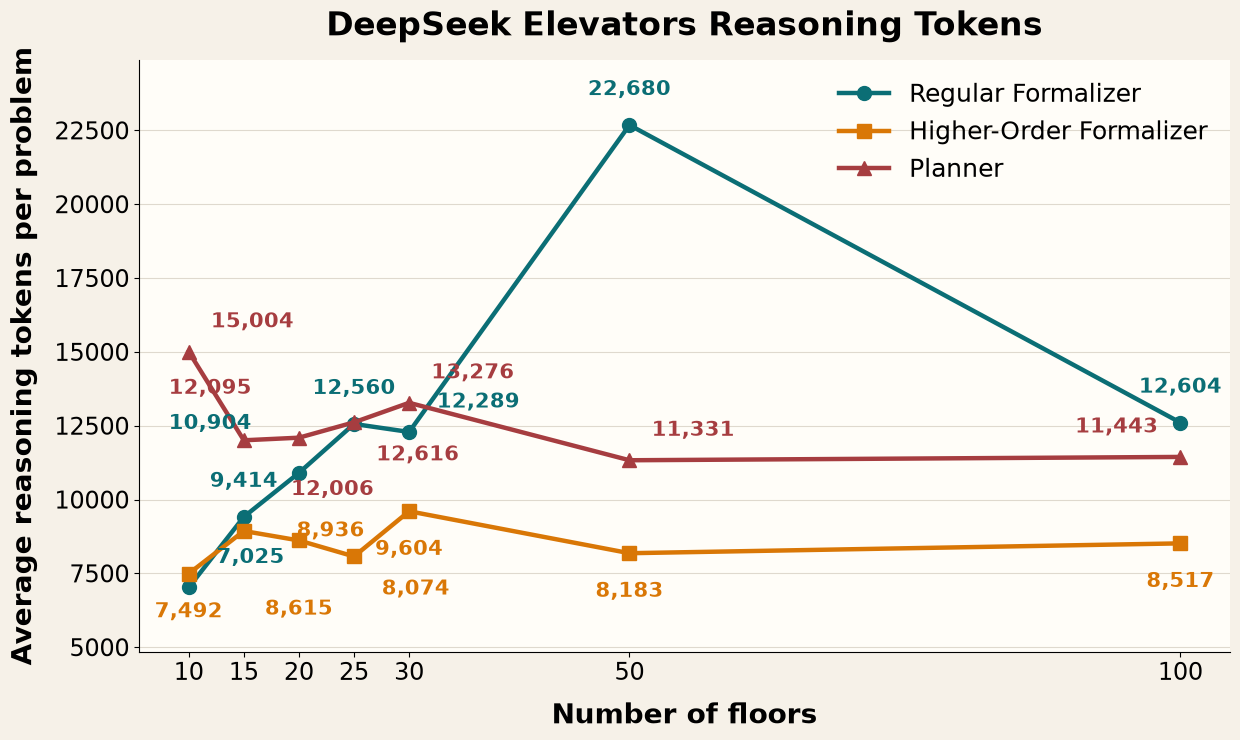}
    \caption{H-O Formalizer significantly reduces reasoning token consumption in Elevators.}
    \label{fig:elevators_reasoning_tokens}
    \vspace{-2ex}
\end{figure}

Figure~\ref{fig:merged_scaling_curves_all}:b-e shows that the unraveling problems in \unravel almost zero-out Planner performance. 
Figure \ref{fig:merged_scaling_curves_all}:b-e also indicates that when comparing Formalizer and H-O Formalizer, \textbf{H-O Formalizer outperforms Formalizer in nearly all domains, especially for more complex problems}.  Why can Formalizer with frontier models keep robust? Is it because of true scalability or because the specific problems are simply too easy? To test this, we extend the problems to 300 and 1000 children in ChildSnack. For the 300-batch in ChildSnack, \texttt{G3F} and \texttt{DS-V4} achieve 80\% and 90\%, respectively. For the 1000 batch, the models are not able to complete the problems. Although performance starts to decline slightly as the problems become extremely large, this brief extension study suggests that the key reason for the observed Formalizer robustness is likely true scalability. Furthermore, we also collect data on the reasoning tokens that are consumed in Planner, Formalizer, and H-O Formalizer. Using \texttt{DS-V4} as an example, \textbf{Figures \ref{fig:childsnack_reasoning_tokens} and \ref{fig:elevators_reasoning_tokens} illustrate that Formalizer can consume 1.5 to nearly 4 times more reasoning tokens than H-O Formalizer} in complex problems of ChildSnack and Elevators. Figures \ref{fig:childsnack_reasoning_tokens} and \ref{fig:elevators_reasoning_tokens} also reveal that the token consumption of H-O Formalizer reasoning is agnostic to problem-size and stays consistently low. \textbf{Overall, our results show that H-O Formalizer outperforms Formalizer in both accuracy and token consumption}. 

\section{Discussion}
To demonstrate the effectiveness of H-O Formalizer, we conduct ablation studies using \texttt{Qwen2.5-Coder-32B-Instruct (\texttt{Q25})} in all four domains in which we do not ask \texttt{Q25} to reflect on repeating patterns code, and we perform error analysis on \texttt{Q25}'s code by domains under the ablated prompting. We randomly picked two batches of problems from two domains and manually analyze at most 10 errors in those batches. 


In Transport, 50\% of errors occur in appending object types, and 50\% occur in either the loop assigning packages to location or the loop assigning capacities to trucks. An example of an object appending error is shown in Listing \ref{lst:Transport_objects_type_error}.

\begin{figure}[t]
    \centering
    \includegraphics[width=0.9\columnwidth]{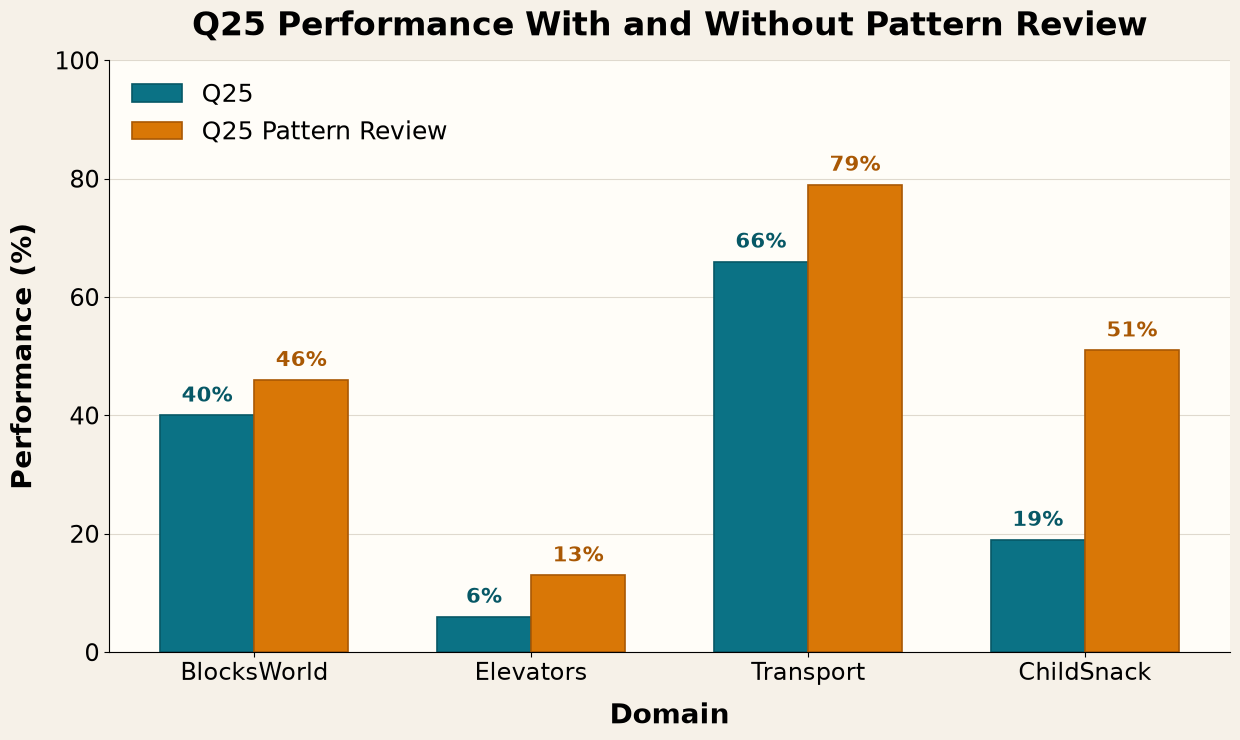}
    \caption{Pattern review boosts \texttt{Q25} performance in all four domains.}
    \label{fig:q25_pattern_review_performance}
    \vspace{-2ex}
\end{figure}

\begin{lstlisting}[language=lisp, caption={Correct vs. Qwen's Code for Appending Objects Types Facts}, label={lst:Transport_objects_type_error}]
Correct code:
lines.append("  (:objects")
lines.append(f"    {' '.join(locations)} - location")
lines.append(f"    {' '.join(vehicles)} - vehicle")
lines.append(f"    {' '.join(packages)} - package")
lines.append(f"    {' '.join(capacities)} - capacity-number")
lines.append("  )")

Qwen writes:
objects = locations + packages + vehicles + [f"capacity-{i}" for i in range(max_capacity + 1)]
\end{lstlisting}
In ChildSnack, 100\% of errors occur in the `waiting' loop assigning children to wait for sandwiches at different places. An example of such an error is shown in Listing \ref{lst:ChildSnack_waiting_error}. The overall performance boost from pattern review is shown in Figure \ref{fig:q25_pattern_review_performance}.

\section{Related Work}
\citet{wei-etal-2025-plangenllms} and \citet{tantakoun-etal-2025-llms} are recent surveys on the Planner and Formalizer methodologies. In essence, LLMs have been tasked to generate plans directly \cite{sel2025llms,verma2025teachingllmsplanlogical,bohnet2025enhancingllmplanningcapabilities} or generate a formal language to interface a symbolic solver \cite{xie2023translating,liu2023llm+,zhang-etal-2024-pddlego,zhang-etal-2024-proc2pddl,zhu2024language}. While Planner is intuitive and strong on small problems, Formalizer promises verifiability and scalability, which we show does not always hold. 
Therefore, we only focus on works involving scaling complexity in planning and beyond.

\paragraph{Scaling Complexity.} Prior work has demonstrated degraded performance as models plan with larger and more complex problems \cite{valmeekam2024llmscantplanlrms, shojaee2025illusionthinkingunderstandingstrengths, lin2025zebralogic, vyas2025are, vyas2025llmswarmpddlempoweredhive}. \citet{10.1007/978-3-032-11402-0_9, shi2023largelanguagemodelseasily} reveal that diminishing performance may be due to large text size and irrelevant context added to the problem. \citet{kagitha2025addressingchallengesplanninglanguage} demonstrates potential scaling ability for Formalizer, while \citet{huang2025languagemodelplannerformalizer} shows that both methodologies struggle when constraints are added to larger problems. \citet{vyas2025llmswarmpddlempoweredhive} also illustrates the degrading performance pattern, particularly in their output category, when approaching planning and task selection from a more agentic perspective. However, no work has been done on unraveling problems, the focus of this paper, except \citet{li2026constructingindustrialscaleoptimizationmodeling} which discusses description compression from an information theoretic angle. Nevertheless, \citet{li2026constructingindustrialscaleoptimizationmodeling} focuses on mixed-integer optimization models, whereas our paper isolates the description-to-grounded-representation expansion bottleneck in classical planning and tests whether generating compact programs instead of explicit PDDL overcomes it.

\paragraph{Code Generation.} Using H-O Formalizer relies on LLMs' ability to generate code. LLMs have been used for writing and debugging programs \cite{Jiang_2026} and generating code as intermediate representations for PDDL formalization \cite{kagitha2025addressingchallengesplanninglanguage}. Related to H-O Formalizer, LLMs have also been used to generate solver code to produce plans \cite{stein2026improvedgeneralizedplanningllms, silver2023generalizedplanningpddldomains}. However, different from prior work, we generate code that will be used to formalize the environment, rather than generate a plan.

\section{Conclusion}
We revisit the scalability of LLM-based planning and argue that prior evaluations overestimate the robustness of Formalizer due to small sizes and unrealistic one-to-one NL-PDDL mappings in existing benchmarks. To address this, we introduce unraveling problems, where compact descriptions expand into large grounded planning instances. Our experiments across four planning domains show that standard Formalizers also degrade under this setting. We further propose H-O Formalizer, which leverages LLMs' code-generation ability to produce compact generator programs instead of explicit grounded representations. This paradigm improves scalability, particularly on large and highly compressed planning problems.

\section{Limitations}
We discuss a few important limitations of our method. First, since we use exact-match parsers to evaluate the differences between generated and ground-truth PDDLs, our accuracy percentages could represent stricter lower-bounds for the potential ranges. Although this is still applied to all models consistently, future work could allow more room for semantic correctness in the difference checking to observe if there are improved performances for complex problems.

Second, although we evaluate four planning domains with different structures and parameterizations, our experiments are still limited to classical symbolic planning settings with fixed domain specifications. Future work may extend unraveling problems to broader planning environments, such as partially observable, stochastic, temporal, or multi-agent domains, where compact natural-language descriptions may induce even more complex grounded representations and execution constraints.

Finally, our implementation of H-O Formalizer relies on Python-based generator programs and handcrafted prompting strategies, including the pattern-reflection stage. Although this design demonstrates strong empirical scalability, it does not yet explore whether the generated higher-order representations are optimal, minimal, or transferable across domains. Future research may investigate more principled intermediate representations, learned compilation strategies, or tighter integration between higher-order formalization and lifted planning systems.

\bibliography{anthology, custom}

\appendix
\section{Formalizer vs. H-O Formalizer Output Tokens Comparison}
\begin{figure}[H]
    \centering

    \begin{subfigure}{0.9\columnwidth}
        \centering
        \includegraphics[width=\linewidth]{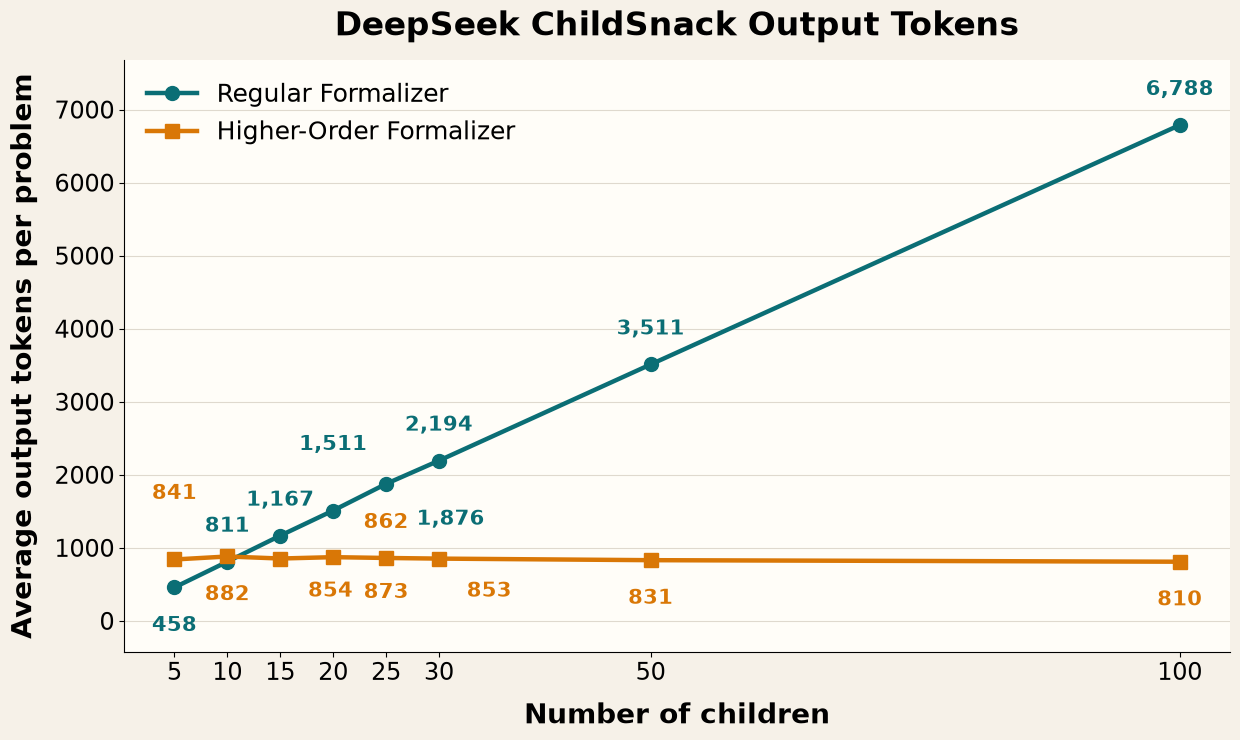}
        \label{fig:deepseek_childsnack_output_tokens}
    \end{subfigure}

    \vspace{1ex}

    \begin{subfigure}{0.9\columnwidth}
        \centering
        \includegraphics[width=\linewidth]{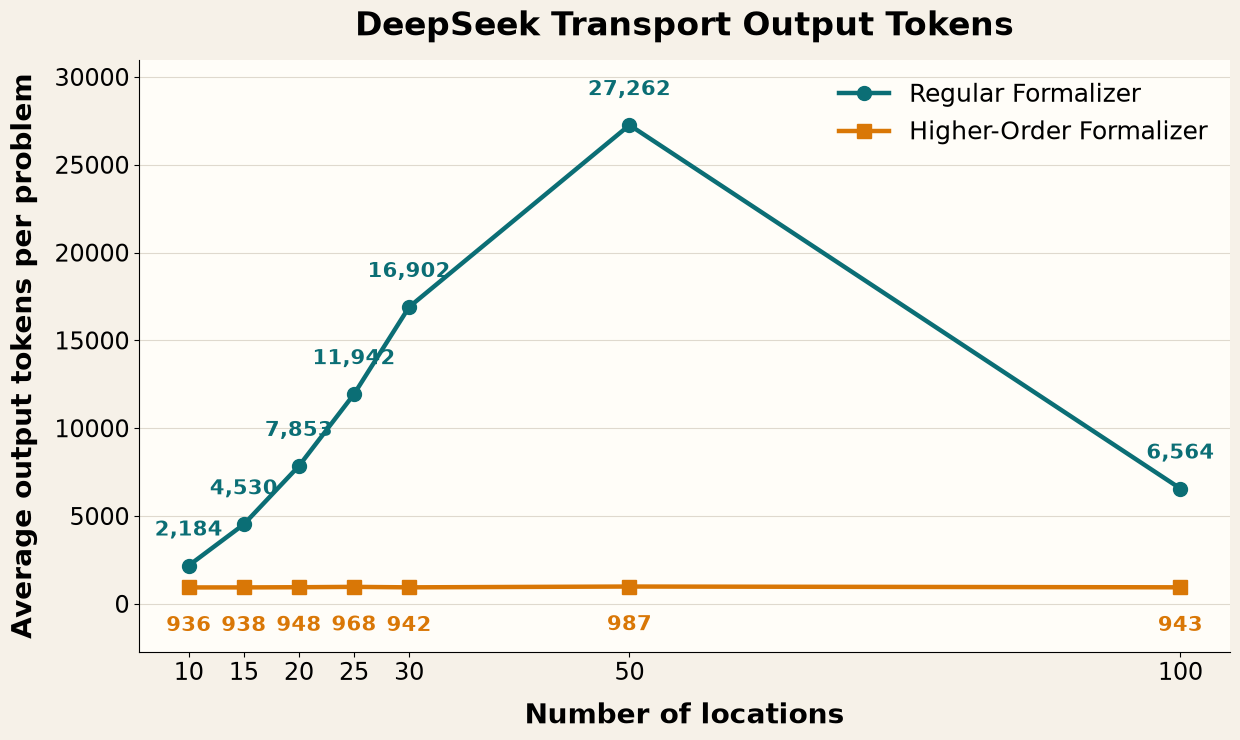}
        \label{fig:deepseek_transport_output_tokens}
    \end{subfigure}

    \vspace{1ex}

    \begin{subfigure}{0.9\columnwidth}
        \centering
        \includegraphics[width=\linewidth]{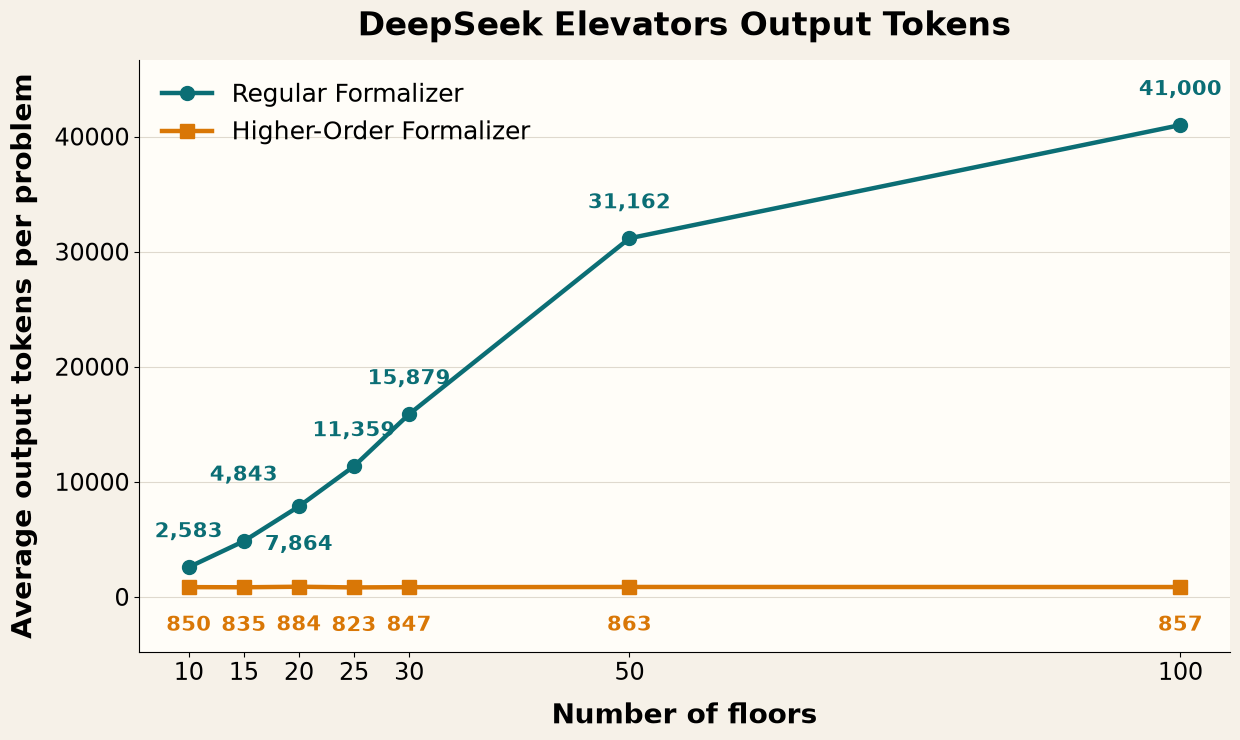}
        \label{fig:deepseek_elevators_output_tokens}
    \end{subfigure}

    \vspace{1ex}

    \begin{subfigure}{0.9\columnwidth}
        \centering
        \includegraphics[width=\linewidth]{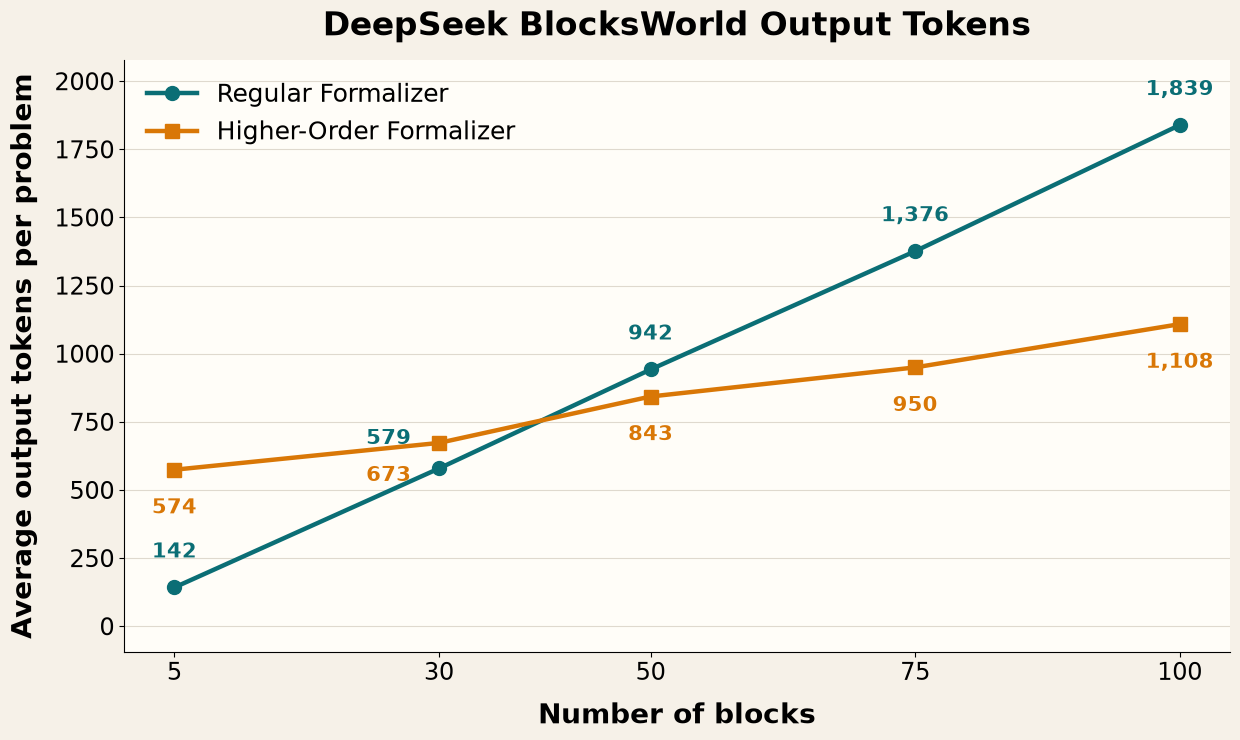}
        \label{fig:deepseek_blocksworld_output_tokens}
    \end{subfigure}

    \caption{Formalizer output token consumption increases as problems become more complex and eventually consumes significantly more output tokens than H-O Formalizer. H-O Formalizer often consumes consistently low output tokens by generating Python scripts. In Transport, the output tokens for the last problem complexity drops because the model starts missing a large chunk of the correct initial states.}
    \label{fig:deepseek_output_tokens}
    \vspace{-2ex}
\end{figure}

\section{Example Errors}
\label{app:example_code_errors}
\begin{lstlisting}[language=lisp, caption={Correct vs. Qwen's Code for `Waiting' Facts}, label={lst:ChildSnack_waiting_error}]
Correct code:
for i, child in enumerate(children, start=1):
    if i % 3 == 0:
        init_facts.append(f"(allergic_gluten {child})")
    else:
        init_facts.append(f"(not_allergic_gluten {child})")

    if child_num == 1 or child_num >= 22:
            table = "table3"
    elif child_num <= 11:
            table = "table1"
    else:
            table = "table2"
    init_facts.append(f"(waiting {child} {tables[table_index]})")

Qwen writes:
for i, child in enumerate(children, start=1):
    if i % 3 == 0:
        init_facts.append(f"(allergic_gluten {child})")
    else:
        init_facts.append(f"(not_allergic_gluten {child})")
    table_index = (i - 1) // children_per_tray
    init_facts.append(f"(waiting {child} {tables[table_index]})")
\end{lstlisting}



\section{Main and Fixed Parameters in Transport and Elevators}

\label{app:main_fixed_parameters}
\begin{table}[H]
\centering
\footnotesize
\setlength{\tabcolsep}{2.5pt}
\renewcommand{\arraystretch}{1.08}
\begin{tabularx}{\columnwidth}{@{}lX X c c@{}}
\toprule
Domain & Varied & Fixed & Ours & Max \\
\midrule
\multirow{3}{*}{Transport}
& \multirow{3}{*}{Locations} & Packages & 10 & 30 \\
& & Vehicles & 3 & 11 \\
& & Capacity & 4 & -- \\
\midrule
\multirow{2}{*}{Elevators}
& \multirow{2}{*}{Floors} & Passengers & 15 & 39 \\
& & Slow Elevator & 2 & 2 \\
& & Fast Elevator & 1 & 3 \\
& & Slow Capacity & 4 & 4 \\
& & Fast Capacity & 6 & 6 \\
\bottomrule
\end{tabularx}
\caption{Main (varied) and fixed parameters in the unraveling problems of Transport and Elevators. Dashes indicate that there are no directly comparable maximum in the original competition settings, and we set the values in our problems to create controlled, structured problem templates.}
\label{tab:unraveling-parameters}
\end{table}

\section{Problem Size and Plan Length Relation} \label{app:plan_length}
To better consider the linear relationship between the natural language inputs and the resulting solving process under PDDL formulation, we plot number of blocks and the resulting plan length (in lines of text file) (\autoref{fig:plan_length_curve}).

\section{BlocksWorld-XXL Dataset Construction} 
\label{app:extension}
Problems were generated following the procedure in \cite{huang-zhang-2025-limit}. For BlocksWorld-XXL, we first randomly generate the initial stacks and goal stacks given the number of blocks, then use a template to automatically generate the problem descriptions. As mentioned in Section \ref{sec: unraveling problems} the initial stacks were artificially set. Listings 4-19 display example PDDL and natural language descriptions for BlocksWorld-XXL, BlocksWorld-Unravel, ChildSnack, Elevators, and Transport.

\begin{figure}[!t]
    \centering
    \pgfplotsset{
        every axis/.append style={
            width=\linewidth,
            height=4cm,
            xmin=0, xmax=100, 
            ymin=0, ymax=200,
            xtick={0,20,40,60,80, 100},
            x tick label style={font=\tiny},
            ytick={0,50,100,150,200,250},
            yticklabel={\pgfmathprintnumber{\tick}},
            tick label style={font=\tiny},
            title style={font=\scriptsize\bfseries, yshift=-1ex},
            grid=major,
            grid style={dashed, gray!30},
        }
    }

    \textbf{\footnotesize BlocksWorld-XXL}\par\vspace{1mm}

    \begin{subfigure}{0.5\columnwidth}
        \begin{tikzpicture}
            \begin{axis}[
                title={Average Plan length}, 
                legend entries={Gemini 3 Flash, Qwen3-32B, Qwen2.5-32B},
                legend to name=sharedlegend,
                legend columns=-1, 
                legend style={font=\scriptsize, /tikz/every even column/.append style={column sep=0.3cm}}
            ]
                \addplot[color=blue, mark=square*, mark size=1.5pt, thick] coordinates {
                    (5, 8) (30, 80) (50, 109) (75, 181)
                };
                \addplot[color=red, mark=triangle*, mark size=1.5pt, thick] coordinates {
                    (5, 8) (30, 68) (50, 108) (75, 116)
                };
                \addplot[color=green!60!black, mark=*, mark size=1.5pt, thick] coordinates {
                    (5, 8) (30, 77) (50, 114) (75, 149)
                };
            \end{axis}
        \end{tikzpicture}
    \end{subfigure}%
    \hspace{-0.2cm}
    
    \vspace{-0.3cm} 
    
    \begin{center}
        \ref{sharedlegend}
    \end{center}

    \caption{Average plan length increases with number of blocks under LLM-as-formalizer pipeline, indicating growing solution complexity with increasing input complexity.}
    \label{fig:plan_length_curve}
\end{figure}

\section{Experimental Details} \label{app:prompts}
During evaluation of formalizers, due to the large problem sizes in our dataset, the solver often crashes before completion, so we implement a parser to compare the differences between the LLM-generated and ground-truth problem files. Knowing our domains' problems are always solvable, we regard a perfect match for all of the objects, initial, and goal states between the two problem files implies a valid plan, and any mismatch means invalid plan. 

Listings 20-33 display the full prompts that we use to generate the outputs, including plans and problem files. 

\section{Potential Risks}
Although LLM-as-formalizer and LLM-as-higher-order-formalizer have much higher accuracy than LLM-as-planner, it is important to note that formalizer methods may still experience issues like hallucination, especially if they are applied in the real world.

\section{Personally Identifying Information}
The data used in this project is synthetic and does not contain personally identifying information or offensive content.

\section{Descriptive Statistics}
All statistics for experimental results are in percentages. Each experiment is run once, as each result only takes extreme values of either 0\% or 100\%.

\section{Artifacts Licenses and Use}
Our use of artifacts is completely for academic research and fit the intended licenses or use purposes.
\onecolumn

\begin{lstlisting}[language=lisp, caption={Domain File for BlocksWorld-XXL and BlocksWorld-Unravel}, label={lst:blocksworld_df}]
(define (domain blocksworld)
  (:requirements :strips)
(:predicates (clear ?x)
             (on-table ?x)
             (arm-empty)
             (holding ?x)
             (on ?x ?y))

(:action pickup
  :parameters (?ob)
  :precondition (and (clear ?ob) (on-table ?ob) (arm-empty))
  :effect (and (holding ?ob) (not (clear ?ob)) (not (on-table ?ob)) 
               (not (arm-empty))))

(:action putdown
  :parameters  (?ob)
  :precondition (holding ?ob)
  :effect (and (clear ?ob) (arm-empty) (on-table ?ob) 
               (not (holding ?ob))))

(:action stack
  :parameters  (?ob ?underob)
  :precondition (and (clear ?underob) (holding ?ob))
  :effect (and (arm-empty) (clear ?ob) (on ?ob ?underob)
               (not (clear ?underob)) (not (holding ?ob))))

(:action unstack
  :parameters  (?ob ?underob)
  :precondition (and (on ?ob ?underob) (clear ?ob) (arm-empty))
  :effect (and (holding ?ob) (clear ?underob)
               (not (on ?ob ?underob)) (not (clear ?ob)) (not (arm-empty)))))   
\end{lstlisting}

\begin{lstlisting}[language=lisp, caption={Domain Description for an Example 100-Block Problem in BlocksWorld-XXL}, label={lst:blocksworld_dd}]
I am playing with a set of blocks where I need to arrange the blocks into stacks. Here are the actions I can do

 Pick up a block
 Unstack a block from on top of another block
 Put down a block
 Stack a block on top of another block

 I have the following restrictions on my actions:
 I can only pick up or unstack one block at a time.
 I can only pick up or unstack a block if my hand is empty.
 I can only pick up a block if the block is on the table and the block is clear. A block is clear if the block has no other blocks on top of it and if the block is not picked up.
 I can only unstack a block from on top of another block if the block I am unstacking was really on top of the other block.
 I can only unstack a block from on top of another block if the block I am unstacking is clear.
 Once I pick up or unstack a block, I am holding the block.
 I can only put down a block that I am holding.
 I can only stack a block on top of another block if I am holding the block being stacked.
 I can only stack a block on top of another block if the block onto which I am stacking the block is clear.
 Once I put down or stack a block, my hand becomes empty.
 Once you stack a block on top of a second block, the second block is no longer clear.
\end{lstlisting}

\begin{lstlisting}[language=lisp, caption={Problem Description for an Example 5-Block Problem in BlocksWorld-XXL}, label={lst:blocksworld_pdsmall}]
As initial conditions I have that, block 1 is clear, the hand is empty, block 1 is on top of block 3, block 3 is on top of block 5, block 4 is on top of block 2, block 5 is on top of block 4, and block 2 is on the table.
My goal is to have that block 1 is on the table, block 2 is on the table, block 3 is on the table, block 4 is on the table, and block 5 is on the table.
\end{lstlisting}

\begin{lstlisting}[
  language=lisp,
  caption={Problem Description for an Example 100-Block Problem in BlocksWorld-XXL},
  label={lst:blocksworld_pdlarge},
  breaklines=true
]
As initial conditions I have that, 
block 1 is clear, block 2 is clear, block 3 is clear, block 4 is clear, 
block 5 is clear, block 6 is clear, block 7 is clear, block 8 is clear, 
block 9 is clear, block 10 is clear, block 11 is clear, 
block 12 is clear, block 14 is clear, block 15 is clear, 
block 16 is clear, block 18 is clear, block 19 is clear, 
block 20 is clear, block 23 is clear, block 24 is clear, 
block 25 is clear, block 26 is clear, block 27 is clear, 
block 28 is clear, block 29 is clear, block 30 is clear, 
block 31 is clear, block 33 is clear, block 36 is clear, 
block 38 is clear, block 39 is clear, block 40 is clear, 
block 41 is clear, block 42 is clear, block 43 is clear, 
block 44 is clear, block 46 is clear, block 47 is clear, 
block 48 is clear, block 50 is clear, block 51 is clear, 
block 52 is clear, block 53 is clear, block 54 is clear, 
block 56 is clear, block 57 is clear, block 58 is clear, 
block 59 is clear, block 60 is clear, block 61 is clear, 
block 62 is clear, block 63 is clear, block 64 is clear, 
block 65 is clear, block 68 is clear, block 69 is clear, 
block 73 is clear, block 74 is clear, block 75 is clear, 
block 76 is clear, block 77 is clear, block 78 is clear, 
block 80 is clear, block 81 is clear, block 83 is clear, 
block 84 is clear, block 85 is clear, block 86 is clear, 
block 87 is clear, block 88 is clear, block 90 is clear, 
block 93 is clear, block 94 is clear, block 95 is clear, 
block 96 is clear, block 97 is clear, block 98 is clear, 
block 100 is clear, the hand is empty, block 8 is on top of block 35, 
block 12 is on top of block 82, block 14 is on top of block 32, 
block 15 is on top of block 92, block 16 is on top of block 72, 
block 18 is on top of block 67, block 19 is on top of block 37, 
block 20 is on top of block 34, block 24 is on top of block 49, 
block 27 is on top of block 13, block 28 is on top of block 66, 
block 32 is on top of block 17, block 34 is on top of block 91, 
block 35 is on top of block 22, block 47 is on top of block 89, 
block 59 is on top of block 79, block 67 is on top of block 71, 
block 69 is on top of block 70, block 79 is on top of block 55, 
block 81 is on top of block 99, block 89 is on top of block 45, 
block 94 is on top of block 21, block 1 is on the table, 
block 2 is on the table, block 3 is on the table, 
block 4 is on the table, block 5 is on the table, 
block 6 is on the table, block 7 is on the table, 
block 9 is on the table, block 10 is on the table, 
block 11 is on the table, block 13 is on the table, 
block 17 is on the table, block 21 is on the table, 
block 22 is on the table, block 23 is on the table, 
block 25 is on the table, block 26 is on the table, 
block 29 is on the table, block 30 is on the table, 
block 31 is on the table, block 33 is on the table, 
block 36 is on the table, block 37 is on the table, 
block 38 is on the table, block 39 is on the table, 
block 40 is on the table, block 41 is on the table, 
block 42 is on the table, block 43 is on the table, 
block 44 is on the table, block 45 is on the table, 
block 46 is on the table, block 48 is on the table, 
block 49 is on the table, block 50 is on the table, 
block 51 is on the table, block 52 is on the table, 
block 53 is on the table, block 54 is on the table, 
block 55 is on the table, block 56 is on the table, 
block 57 is on the table, block 58 is on the table, 
block 60 is on the table, block 61 is on the table, 
block 62 is on the table, block 63 is on the table, 
block 64 is on the table, block 65 is on the table, 
block 66 is on the table, block 68 is on the table, 
block 70 is on the table, block 71 is on the table, 
block 72 is on the table, block 73 is on the table, 
block 74 is on the table, block 75 is on the table, 
block 76 is on the table, block 77 is on the table, 
block 78 is on the table, block 80 is on the table, 
block 82 is on the table, block 83 is on the table, 
block 84 is on the table, block 85 is on the table, 
block 86 is on the table, block 87 is on the table, 
block 88 is on the table, block 90 is on the table, 
block 91 is on the table, block 92 is on the table, 
block 93 is on the table, block 95 is on the table, 
block 96 is on the table, block 97 is on the table, 
block 98 is on the table, block 99 is on the table, 
and block 100 is on the table.
My goal is to have that,
block 2 is on top of block 39, block 8 is on top of block 42, 
block 10 is on top of block 92, block 12 is on top of block 10, 
block 15 is on top of block 56, block 16 is on top of block 14, 
block 19 is on top of block 20, block 20 is on top of block 25, 
block 31 is on top of block 64, block 32 is on top of block 90, 
block 33 is on top of block 67, block 40 is on top of block 62, 
block 41 is on top of block 22, block 42 is on top of block 60, 
block 43 is on top of block 12, block 46 is on top of block 3, 
block 55 is on top of block 57, block 57 is on top of block 71, 
block 58 is on top of block 63, block 62 is on top of block 4, 
block 66 is on top of block 36, block 68 is on top of block 74, 
block 69 is on top of block 82, block 70 is on top of block 34, 
block 71 is on top of block 77, block 73 is on top of block 27, 
block 78 is on top of block 21, block 79 is on top of block 44, 
block 80 is on top of block 58, block 81 is on top of block 91, 
block 82 is on top of block 7, block 83 is on top of block 18, 
block 84 is on top of block 68, block 86 is on top of block 40, 
block 88 is on top of block 1, block 91 is on top of block 48, 
block 93 is on top of block 51, block 1 is on the table, 
block 3 is on the table, block 4 is on the table, 
block 5 is on the table, block 6 is on the table, 
block 7 is on the table, block 9 is on the table, 
block 11 is on the table, block 13 is on the table, 
block 14 is on the table, block 17 is on the table,
block 18 is on the table, block 21 is on the table, 
block 22 is on the table, block 23 is on the table, 
block 24 is on the table, block 25 is on the table, 
block 26 is on the table, block 27 is on the table, 
block 28 is on the table, block 29 is on the table, 
block 30 is on the table, block 34 is on the table, 
block 35 is on the table, block 36 is on the table, 
block 37 is on the table, block 38 is on the table, 
block 39 is on the table, block 44 is on the table, 
block 45 is on the table, block 47 is on the table, 
block 48 is on the table, block 49 is on the table, 
block 50 is on the table, block 51 is on the table, 
block 52 is on the table, block 53 is on the table, 
block 54 is on the table, block 56 is on the table, 
block 59 is on the table, block 60 is on the table, 
block 61 is on the table, block 63 is on the table, 
block 64 is on the table, block 65 is on the table, 
block 67 is on the table, block 72 is on the table, 
block 74 is on the table, block 75 is on the table, 
block 76 is on the table, block 77 is on the table, 
block 85 is on the table, block 87 is on the table, 
block 89 is on the table, block 90 is on the table, 
block 92 is on the table, block 94 is on the table, 
block 95 is on the table, block 96 is on the table, 
block 97 is on the table, block 98 is on the table, 
block 99 is on the table, and block 100 is on the table.
\end{lstlisting}

\begin{lstlisting}[language=lisp, caption={Problem Description for an Example 100-Block Problem in BlocksWorld-Unravel}, label={lst:blocksworld_pdunravel}]
As initial conditions I have 100 blocks numbered 1 to 100. All the odd numbered blocks are on one stack, with increasing numbering from top to bottom. Same for the even numbered blocks on another stack.
My goal is to have that block 1 is on top of block 74, block 2 is on top of block 47, block 3 is on top of block 36, block 4 is on top of block 6, block 5 is on top of block 93, block 6 is on top of block 9, block 7 is on top of block 1, block 8 is on top of block 83, block 9 is on top of block 77, block 10 is on top of block 44, block 11 is on top of block 34, block 12 is on top of block 63, block 14 is on top of block 8, block 15 is on top of block 65, block 16 is on top of block 14, block 17 is on top of block 92, block 18 is on top of block 32, block 19 is on top of block 7, block 20 is on top of block 26, block 21 is on top of block 84, block 23 is on top of block 40, block 24 is on top of block 37, block 25 is on top of block 15, block 26 is on top of block 48, block 27 is on top of block 95, block 28 is on top of block 59, block 29 is on top of block 35, block 30 is on top of block 90, block 31 is on top of block 17, block 32 is on top of block 55, block 33 is on top of block 91, block 34 is on top of block 85, block 35 is on top of block 46, block 36 is on top of block 97, block 37 is on top of block 11, block 38 is on top of block 82, block 39 is on top of block 71, block 40 is on top of block 81, block 41 is on top of block 75, block 42 is on top of block 73, block 43 is on top of block 87, block 44 is on top of block 22, block 45 is on top of block 66, block 46 is on top of block 80, block 47 is on top of block 96, block 48 is on top of block 50, block 49 is on top of block 89, block 50 is on top of block 4, block 51 is on top of block 21, block 52 is on top of block 99, block 53 is on top of block 64, block 54 is on top of block 88, block 55 is on top of block 25, block 56 is on top of block 45, block 57 is on top of block 62, block 58 is on top of block 67, block 59 is on top of block 98, block 60 is on top of block 18, block 61 is on top of block 20, block 62 is on top of block 100, block 63 is on top of block 43, block 64 is on top of block 31, block 65 is on top of block 27, block 66 is on top of block 12, block 67 is on top of block 42, block 68 is on top of block 94, block 69 is on top of block 41, block 71 is on top of block 10, block 72 is on top of block 33, block 73 is on top of block 79, block 74 is on top of block 53, block 75 is on top of block 39, block 76 is on top of block 58, block 77 is on top of block 78, block 78 is on top of block 38, block 79 is on top of block 13, block 80 is on top of block 57, block 81 is on top of block 61, block 82 is on top of block 24, block 83 is on top of block 56, block 84 is on top of block 5, block 85 is on top of block 69, block 86 is on top of block 52, block 87 is on top of block 70, block 88 is on top of block 29, block 89 is on top of block 16, block 90 is on top of block 72, block 91 is on top of block 28, block 92 is on top of block 51, block 93 is on top of block 3, block 94 is on top of block 60, block 95 is on top of block 54, block 97 is on top of block 23, block 98 is on top of block 49, block 99 is on top of block 19, block 100 is on top of block 30, block 13 is on the table, block 22 is on the table, block 70 is on the table, and block 96 is on the table.
\end{lstlisting}

\begin{lstlisting}[language=lisp, caption={Domain File for ChildSnack}, label={lst:childsnack_df}]
(define (domain child-snack)
(:requirements :typing :equality)
(:types child bread-portion content-portion sandwich tray place)
(:constants kitchen - place)

(:predicates (at_kitchen_bread ?b - bread-portion)
	     (at_kitchen_content ?c - content-portion)
     	     (at_kitchen_sandwich ?s - sandwich)
     	     (no_gluten_bread ?b - bread-portion)
       	     (no_gluten_content ?c - content-portion)
      	     (ontray ?s - sandwich ?t - tray)
       	     (no_gluten_sandwich ?s - sandwich)
	     (allergic_gluten ?c - child)
     	     (not_allergic_gluten ?c - child)
	     (served ?c - child)
	     (waiting ?c - child ?p - place)
             (at ?t - tray ?p - place)
	     (notexist ?s - sandwich)
  )

(:action make_sandwich_no_gluten
	 :parameters (?s - sandwich ?b - bread-portion ?c - content-portion)
	 :precondition (and (at_kitchen_bread ?b)
			    (at_kitchen_content ?c)
			    (no_gluten_bread ?b)
			    (no_gluten_content ?c)
			    (notexist ?s))
	 :effect (and
		   (not (at_kitchen_bread ?b))
		   (not (at_kitchen_content ?c))
		   (at_kitchen_sandwich ?s)
		   (no_gluten_sandwich ?s)
                   (not (notexist ?s))
		   ))


(:action make_sandwich
	 :parameters (?s - sandwich ?b - bread-portion ?c - content-portion)
	 :precondition (and (at_kitchen_bread ?b)
			    (at_kitchen_content ?c)
                            (notexist ?s)
			    )
	 :effect (and
		   (not (at_kitchen_bread ?b))
		   (not (at_kitchen_content ?c))
		   (at_kitchen_sandwich ?s)
                   (not (notexist ?s))
		   ))


(:action put_on_tray
	 :parameters (?s - sandwich ?t - tray)
	 :precondition (and  (at_kitchen_sandwich ?s)
			     (at ?t kitchen))
	 :effect (and
		   (not (at_kitchen_sandwich ?s))
		   (ontray ?s ?t)))


(:action serve_sandwich_no_gluten
 	:parameters (?s - sandwich ?c - child ?t - tray ?p - place)
	:precondition (and
		       (allergic_gluten ?c)
		       (ontray ?s ?t)
		       (waiting ?c ?p)
		       (no_gluten_sandwich ?s)
                       (at ?t ?p)
		       )
	:effect (and (not (ontray ?s ?t))
		     (served ?c)))

(:action serve_sandwich
	:parameters (?s - sandwich ?c - child ?t - tray ?p - place)
	:precondition (and (not_allergic_gluten ?c)
	                   (waiting ?c ?p)
			   (ontray ?s ?t)
			   (at ?t ?p))
	:effect (and (not (ontray ?s ?t))
		     (served ?c)))

(:action move_tray
	 :parameters (?t - tray ?p1 ?p2 - place)
	 :precondition (and (at ?t ?p1))
	 :effect (and (not (at ?t ?p1))
		      (at ?t ?p2)))


)
\end{lstlisting}

\begin{lstlisting}[language=lisp, caption={Example Problem File for ChildSnack}, label={lst:childsnack_pf}]
(define (problem p01)
  (:domain child-snack)
  (:objects
    child1 child2 child3 child4 child5 - child
    bread1 bread2 bread3 bread4 bread5 - bread-portion
    content1 content2 content3 content4 content5 - content-portion
    tray1 tray2 tray3 - tray
    table1 table2 table3 - place
    sandw1 sandw2 sandw3 sandw4 sandw5 - sandwich
  )
  (:init
    (at tray1 kitchen)
    (at tray2 kitchen)
    (at tray3 kitchen)
    (at_kitchen_bread bread1)
    (at_kitchen_bread bread2)
    (at_kitchen_bread bread3)
    (at_kitchen_bread bread4)
    (at_kitchen_bread bread5)
    (at_kitchen_content content1)
    (at_kitchen_content content2)
    (at_kitchen_content content3)
    (at_kitchen_content content4)
    (at_kitchen_content content5)
    (no_gluten_bread bread1)
    (no_gluten_content content1)
    (allergic_gluten child1)
    (not_allergic_gluten child2)
    (not_allergic_gluten child3)
    (no_gluten_bread bread4)
    (no_gluten_content content4)
    (allergic_gluten child4)
    (not_allergic_gluten child5)
    (waiting child1 table1)
    (waiting child2 table1)
    (waiting child3 table2)
    (waiting child4 table2)
    (waiting child5 table3)
    (notexist sandw1)
    (notexist sandw2)
    (notexist sandw3)
    (notexist sandw4)
    (notexist sandw5)
  )
  (:goal
    (and
      (served child1)
      (served child2)
      (served child3)
      (served child4)
      (served child5)
    )
  )
)
\end{lstlisting}

\begin{lstlisting}[language=lisp, caption={Example Domain Description for ChildSnack}, label={lst:childsnack_dd}]
    I am preparing and serving sandwiches to children. Here are the actions I can do

Make a sandwich
Make a gluten-free sandwich
Put a sandwich on a tray
Move a tray from one place to another
Serve a sandwich to a child
Serve a gluten-free sandwich to a child

I have the following restrictions on my actions:
I can only make a sandwich if the required bread portion and content portion are both available in the kitchen.
I can only make a gluten-free sandwich if the bread portion and content portion are both gluten-free and available in the kitchen.
I can only use a sandwich object that does not already exist.
Once I make a sandwich, the corresponding bread and content portions are no longer available in the kitchen.
I can only put a sandwich on a tray if the sandwich is in the kitchen and the tray is in the kitchen.
I can only serve a sandwich to a child if the tray holding the sandwich is at the same place where the child is waiting.
I can only serve a regular sandwich to a child who is not allergic to gluten.
I can only serve a gluten-free sandwich to a child who is allergic to gluten.
I can move a tray from one place to another at any time.
\end{lstlisting}

\begin{lstlisting}[language=lisp, caption={Example Problem Description for ChildSnack}, label={lst:childsnack_pd}]
    As initial conditions I have 20 children numbered 1 to 20, 3 trays numbered 1 to 3, 20 bread portions numbered 1 to 20, 20 content portions numbered 1 to 20, and 20 sandwich placeholders numbered 1 to 20. The children assigned to tray 1 are child 3 to 9. The children assigned to tray 2 are child 10 to 16, and the remaining children are assigned to tray 3. Every 3rd child starting from child 1 must receive a gluten-free sandwich, and all other children can receive a regular sandwich. The bread portions whose numbers match the children that need gluten-free sandwiches are gluten-free, and the content portions whose numbers match the children that need gluten-free sandwiches are gluten-free. All trays start in the kitchen, all bread portions and content portions start in the kitchen, and all sandwiches still need to be prepared and served.
My goal is to have each of the 20 children be served a sandwich. For example, child 1 is served a sandwich, child 2 is served a sandwich, child 3 is served a sandwich, and so on.
\end{lstlisting}

\begin{lstlisting}[language=lisp, caption={Domain File for Elevators}, label={lst:elevators_df}]
    (define (domain elevators-sequencedstrips)
  (:requirements :typing :action-costs)
  (:types   elevator - object
            slow-elevator fast-elevator - elevator
            passenger - object
            count - object
         )

(:predicates
    (passenger-at ?person - passenger ?floor - count)
    (boarded ?person - passenger ?lift - elevator)
    (lift-at ?lift - elevator ?floor - count)
    (reachable-floor ?lift - elevator ?floor - count)
    (above ?floor1 - count ?floor2 - count)
    (passengers ?lift - elevator ?n - count)
    (can-hold ?lift - elevator ?n - count)
    (next ?n1 - count ?n2 - count)
)

(:functions (total-cost) - number
            (travel-slow ?f1 - count ?f2 - count) - number
            (travel-fast ?f1 - count ?f2 - count) - number
)

(:action move-up-slow
  :parameters (?lift - slow-elevator ?f1 - count ?f2 - count )
  :precondition (and (lift-at ?lift ?f1) (above ?f1 ?f2 ) (reachable-floor ?lift ?f2) )
  :effect (and (lift-at ?lift ?f2) (not (lift-at ?lift ?f1)) (increase (total-cost) (travel-slow ?f1 ?f2))))

(:action move-down-slow
  :parameters (?lift - slow-elevator ?f1 - count ?f2 - count )
  :precondition (and (lift-at ?lift ?f1) (above ?f2 ?f1 ) (reachable-floor ?lift ?f2) )
  :effect (and (lift-at ?lift ?f2) (not (lift-at ?lift ?f1)) (increase (total-cost) (travel-slow ?f2 ?f1))))

(:action move-up-fast
  :parameters (?lift - fast-elevator ?f1 - count ?f2 - count )
  :precondition (and (lift-at ?lift ?f1) (above ?f1 ?f2 ) (reachable-floor ?lift ?f2) )
  :effect (and (lift-at ?lift ?f2) (not (lift-at ?lift ?f1)) (increase (total-cost) (travel-fast ?f1 ?f2))))

(:action move-down-fast
  :parameters (?lift - fast-elevator ?f1 - count ?f2 - count )
  :precondition (and (lift-at ?lift ?f1) (above ?f2 ?f1 ) (reachable-floor ?lift ?f2) )
  :effect (and (lift-at ?lift ?f2) (not (lift-at ?lift ?f1)) (increase (total-cost) (travel-fast ?f2 ?f1))))

(:action board
  :parameters (?p - passenger ?lift - elevator ?f - count ?n1 - count ?n2 - count)
  :precondition (and  (lift-at ?lift ?f) (passenger-at ?p ?f) (passengers ?lift ?n1) (next ?n1 ?n2) (can-hold ?lift ?n2) )
  :effect (and (not (passenger-at ?p ?f)) (boarded ?p ?lift) (not (passengers ?lift ?n1)) (passengers ?lift ?n2) ))

(:action leave
  :parameters (?p - passenger ?lift - elevator ?f - count ?n1 - count ?n2 - count)
  :precondition (and  (lift-at ?lift ?f) (boarded ?p ?lift) (passengers ?lift ?n1) (next ?n2 ?n1) )
  :effect (and (passenger-at ?p ?f) (not (boarded ?p ?lift)) (not (passengers ?lift ?n1)) (passengers ?lift ?n2) ))

)
\end{lstlisting}

\begin{lstlisting}[language=lisp, caption={Example Problem File for Elevators}, label={lst:elevators_pf}]
    (define (problem p01)
(:domain elevators-sequencedstrips)
    (:objects
       n0 n1 n2 n3 n4 n5 n6 n7 n8 n9 - count
       p1 p2 p3 p4 p5 p6 p7 p8 p9 p10 p11 p12 p13 p14 p15 - passenger
       fast1 - fast-elevator
       slow1 slow2 - slow-elevator
    )

    (:init
       (next n0 n1)
       (next n1 n2)
       (next n2 n3)
       (next n3 n4)
       (next n4 n5)
       (next n5 n6)
       (next n6 n7)
       (next n7 n8)
       (next n8 n9)

       (above n0 n1)
       (above n0 n2)
       (above n0 n3)
       (above n0 n4)
       (above n0 n5)
       (above n0 n6)
       (above n0 n7)
       (above n0 n8)
       (above n0 n9)
       (above n1 n2)
       (above n1 n3)
       (above n1 n4)
       (above n1 n5)
       (above n1 n6)
       (above n1 n7)
       (above n1 n8)
       (above n1 n9)
       (above n2 n3)
       (above n2 n4)
       (above n2 n5)
       (above n2 n6)
       (above n2 n7)
       (above n2 n8)
       (above n2 n9)
       (above n3 n4)
       (above n3 n5)
       (above n3 n6)
       (above n3 n7)
       (above n3 n8)
       (above n3 n9)
       (above n4 n5)
       (above n4 n6)
       (above n4 n7)
       (above n4 n8)
       (above n4 n9)
       (above n5 n6)
       (above n5 n7)
       (above n5 n8)
       (above n5 n9)
       (above n6 n7)
       (above n6 n8)
       (above n6 n9)
       (above n7 n8)
       (above n7 n9)
       (above n8 n9)

       (lift-at slow1 n0)
       (passengers slow1 n0)
       (can-hold slow1 n1)
       (can-hold slow1 n2)
       (can-hold slow1 n3)
       (can-hold slow1 n4)
       (reachable-floor slow1 n0)
       (reachable-floor slow1 n1)
       (reachable-floor slow1 n2)
       (reachable-floor slow1 n3)
       (reachable-floor slow1 n4)
       (reachable-floor slow1 n5)
       (reachable-floor slow1 n6)
       (reachable-floor slow1 n7)
       (reachable-floor slow1 n8)
       (reachable-floor slow1 n9)
       (lift-at slow2 n0)
       (passengers slow2 n0)
       (can-hold slow2 n1)
       (can-hold slow2 n2)
       (can-hold slow2 n3)
       (can-hold slow2 n4)
       (reachable-floor slow2 n0)
       (reachable-floor slow2 n1)
       (reachable-floor slow2 n2)
       (reachable-floor slow2 n3)
       (reachable-floor slow2 n4)
       (reachable-floor slow2 n5)
       (reachable-floor slow2 n6)
       (reachable-floor slow2 n7)
       (reachable-floor slow2 n8)
       (reachable-floor slow2 n9)
       (lift-at fast1 n0)
       (passengers fast1 n0)
       (can-hold fast1 n1)
       (can-hold fast1 n2)
       (can-hold fast1 n3)
       (can-hold fast1 n4)
       (can-hold fast1 n5)
       (can-hold fast1 n6)
       (reachable-floor fast1 n0)
       (reachable-floor fast1 n1)
       (reachable-floor fast1 n2)
       (reachable-floor fast1 n3)
       (reachable-floor fast1 n4)
       (reachable-floor fast1 n5)
       (reachable-floor fast1 n6)
       (reachable-floor fast1 n7)
       (reachable-floor fast1 n8)
       (reachable-floor fast1 n9)

       (passenger-at p1 n0)
       (passenger-at p2 n0)
       (passenger-at p3 n0)
       (passenger-at p4 n0)
       (passenger-at p5 n0)
       (passenger-at p6 n0)
       (passenger-at p7 n0)
       (passenger-at p8 n0)
       (passenger-at p9 n0)
       (passenger-at p10 n0)
       (passenger-at p11 n0)
       (passenger-at p12 n0)
       (passenger-at p13 n0)
       (passenger-at p14 n0)
       (passenger-at p15 n0)

       (= (travel-slow n0 n1) 6)
       (= (travel-fast n0 n1) 4)
       (= (travel-slow n0 n2) 7)
       (= (travel-fast n0 n2) 7)
       (= (travel-slow n0 n3) 8)
       (= (travel-fast n0 n3) 10)
       (= (travel-slow n0 n4) 9)
       (= (travel-fast n0 n4) 13)
       (= (travel-slow n0 n5) 10)
       (= (travel-fast n0 n5) 16)
       (= (travel-slow n0 n6) 11)
       (= (travel-fast n0 n6) 19)
       (= (travel-slow n0 n7) 12)
       (= (travel-fast n0 n7) 22)
       (= (travel-slow n0 n8) 13)
       (= (travel-fast n0 n8) 25)
       (= (travel-slow n0 n9) 14)
       (= (travel-fast n0 n9) 28)
       (= (travel-slow n1 n2) 6)
       (= (travel-fast n1 n2) 4)
       (= (travel-slow n1 n3) 7)
       (= (travel-fast n1 n3) 7)
       (= (travel-slow n1 n4) 8)
       (= (travel-fast n1 n4) 10)
       (= (travel-slow n1 n5) 9)
       (= (travel-fast n1 n5) 13)
       (= (travel-slow n1 n6) 10)
       (= (travel-fast n1 n6) 16)
       (= (travel-slow n1 n7) 11)
       (= (travel-fast n1 n7) 19)
       (= (travel-slow n1 n8) 12)
       (= (travel-fast n1 n8) 22)
       (= (travel-slow n1 n9) 13)
       (= (travel-fast n1 n9) 25)
       (= (travel-slow n2 n3) 6)
       (= (travel-fast n2 n3) 4)
       (= (travel-slow n2 n4) 7)
       (= (travel-fast n2 n4) 7)
       (= (travel-slow n2 n5) 8)
       (= (travel-fast n2 n5) 10)
       (= (travel-slow n2 n6) 9)
       (= (travel-fast n2 n6) 13)
       (= (travel-slow n2 n7) 10)
       (= (travel-fast n2 n7) 16)
       (= (travel-slow n2 n8) 11)
       (= (travel-fast n2 n8) 19)
       (= (travel-slow n2 n9) 12)
       (= (travel-fast n2 n9) 22)
       (= (travel-slow n3 n4) 6)
       (= (travel-fast n3 n4) 4)
       (= (travel-slow n3 n5) 7)
       (= (travel-fast n3 n5) 7)
       (= (travel-slow n3 n6) 8)
       (= (travel-fast n3 n6) 10)
       (= (travel-slow n3 n7) 9)
       (= (travel-fast n3 n7) 13)
       (= (travel-slow n3 n8) 10)
       (= (travel-fast n3 n8) 16)
       (= (travel-slow n3 n9) 11)
       (= (travel-fast n3 n9) 19)
       (= (travel-slow n4 n5) 6)
       (= (travel-fast n4 n5) 4)
       (= (travel-slow n4 n6) 7)
       (= (travel-fast n4 n6) 7)
       (= (travel-slow n4 n7) 8)
       (= (travel-fast n4 n7) 10)
       (= (travel-slow n4 n8) 9)
       (= (travel-fast n4 n8) 13)
       (= (travel-slow n4 n9) 10)
       (= (travel-fast n4 n9) 16)
       (= (travel-slow n5 n6) 6)
       (= (travel-fast n5 n6) 4)
       (= (travel-slow n5 n7) 7)
       (= (travel-fast n5 n7) 7)
       (= (travel-slow n5 n8) 8)
       (= (travel-fast n5 n8) 10)
       (= (travel-slow n5 n9) 9)
       (= (travel-fast n5 n9) 13)
       (= (travel-slow n6 n7) 6)
       (= (travel-fast n6 n7) 4)
       (= (travel-slow n6 n8) 7)
       (= (travel-fast n6 n8) 7)
       (= (travel-slow n6 n9) 8)
       (= (travel-fast n6 n9) 10)
       (= (travel-slow n7 n8) 6)
       (= (travel-fast n7 n8) 4)
       (= (travel-slow n7 n9) 7)
       (= (travel-fast n7 n9) 7)
       (= (travel-slow n8 n9) 6)
       (= (travel-fast n8 n9) 4)

       (= (total-cost) 0)
    )

    (:goal (and
       (passenger-at p1 n9)
       (passenger-at p2 n6)
       (passenger-at p3 n5)
       (passenger-at p4 n0)
       (passenger-at p5 n2)
       (passenger-at p6 n4)
       (passenger-at p7 n8)
       (passenger-at p8 n7)
       (passenger-at p9 n3)
       (passenger-at p10 n1)
       (passenger-at p11 n8)
       (passenger-at p12 n6)
       (passenger-at p13 n5)
       (passenger-at p14 n9)
       (passenger-at p15 n7)
    ))

    (:metric minimize (total-cost))
)
\end{lstlisting}

\begin{lstlisting}[language=lisp, caption={Example Domain Description for Elevators}, label={lst:elevators_dd}]
    I am moving passengers through a building using elevators. Here are the actions I can do

Move a slow elevator up or down
Move a fast elevator up or down
Board a passenger onto an elevator
Let a passenger leave an elevator

I have the following restrictions on my actions:
I can only move an elevator to a floor that the elevator can reach.
I can only move an elevator up or down between floors that are in the correct above or below relation.
I can only board a passenger if the passenger and the elevator are on the same floor.
I can only board a passenger if the elevator has enough remaining capacity.
Once a passenger boards an elevator, that passenger is no longer waiting on the floor.
I can only let a passenger leave an elevator if that passenger is currently inside the elevator.
Once a passenger leaves an elevator, that passenger is on the elevator's current floor.
Moving elevators has a travel cost that depends on the elevator type and the floors involved.

I have the following travel cost rules:
For any two floors where the destination is above the source by d floors, moving a slow elevator between those floors has travel cost d + 5, and moving a fast elevator between those floors has travel cost 3d + 1.
\end{lstlisting}

\begin{lstlisting}[language=lisp, caption={Example Problem Description for Elevators}, label={lst:elevators_pd}]
    As initial conditions I have 15 passengers numbered 1 to 15, 10 floors numbered 1 to 10, 2 slow elevators numbered 1 to 2, and 1 fast elevator numbered 1. Every passenger starts on floor 1. All elevators start on floor 1. Both slow elevators can visit every floor, and the fast elevator can also visit every floor. Each slow elevator has capacity 4, and the fast elevator has capacity 6. All passengers are initially waiting at floor 1.
My goal is to have that passenger 1 is at destination floor 10, passenger 2 is at destination floor 7, passenger 3 is at destination floor 6, passenger 4 is at destination floor 1, passenger 5 is at destination floor 3, passenger 6 is at destination floor 5, passenger 7 is at destination floor 9, passenger 8 is at destination floor 8, passenger 9 is at destination floor 4, passenger 10 is at destination floor 2, passenger 11 is at destination floor 9, passenger 12 is at destination floor 7, passenger 13 is at destination floor 6, passenger 14 is at destination floor 10, and passenger 15 is at destination floor 8.
\end{lstlisting}

\begin{lstlisting}[language=lisp, caption={Domain File for Transport}, label={lst:transport_df}]
;; Transport sequential
;;

(define (domain transport)
  (:requirements :typing :action-costs)
  (:types
        location target locatable - object
        vehicle package - locatable
        capacity-number - object
  )

  (:predicates 
     (road ?l1 ?l2 - location)
     (at ?x - locatable ?v - location)
     (in ?x - package ?v - vehicle)
     (capacity ?v - vehicle ?s1 - capacity-number)
     (capacity-predecessor ?s1 ?s2 - capacity-number)
  )

  (:functions
     (road-length ?l1 ?l2 - location) - number
     (total-cost) - number
  )

  (:action drive
    :parameters (?v - vehicle ?l1 ?l2 - location)
    :precondition (and
        (at ?v ?l1)
        (road ?l1 ?l2)
      )
    :effect (and
        (not (at ?v ?l1))
        (at ?v ?l2)
        (increase (total-cost) (road-length ?l1 ?l2))
      )
  )

 (:action pick-up
    :parameters (?v - vehicle ?l - location ?p - package ?s1 ?s2 - capacity-number)
    :precondition (and
        (at ?v ?l)
        (at ?p ?l)
        (capacity-predecessor ?s1 ?s2)
        (capacity ?v ?s2)
      )
    :effect (and
        (not (at ?p ?l))
        (in ?p ?v)
        (capacity ?v ?s1)
        (not (capacity ?v ?s2))
        (increase (total-cost) 1)
      )
  )

  (:action drop
    :parameters (?v - vehicle ?l - location ?p - package ?s1 ?s2 - capacity-number)
    :precondition (and
        (at ?v ?l)
        (in ?p ?v)
        (capacity-predecessor ?s1 ?s2)
        (capacity ?v ?s1)
      )
    :effect (and
        (not (in ?p ?v))
        (at ?p ?l)
        (capacity ?v ?s2)
        (not (capacity ?v ?s1))
        (increase (total-cost) 1)
      )
  )

)
\end{lstlisting}

\begin{lstlisting}[language=lisp, caption={Example Problem File for Transport}, label={lst:transport_pf}]
    (define (problem p01)
 (:domain transport)
 (:objects
  location-1 - location
  location-2 - location
  location-3 - location
  location-4 - location
  location-5 - location
  location-6 - location
  location-7 - location
  location-8 - location
  location-9 - location
  location-10 - location
  truck-1 - vehicle
  truck-2 - vehicle
  truck-3 - vehicle
  package-1 - package
  package-2 - package
  package-3 - package
  package-4 - package
  package-5 - package
  package-6 - package
  package-7 - package
  package-8 - package
  package-9 - package
  package-10 - package
  capacity-0 - capacity-number
  capacity-1 - capacity-number
  capacity-2 - capacity-number
  capacity-3 - capacity-number
  capacity-4 - capacity-number
 )
 (:init
  (= (total-cost) 0)
  (capacity-predecessor capacity-0 capacity-1)
  (capacity-predecessor capacity-1 capacity-2)
  (capacity-predecessor capacity-2 capacity-3)
  (capacity-predecessor capacity-3 capacity-4)
  (road location-1 location-2)
  (= (road-length location-1 location-2) 1)
  (road location-1 location-3)
  (= (road-length location-1 location-3) 1)
  (road location-1 location-4)
  (= (road-length location-1 location-4) 1)
  (road location-1 location-5)
  (= (road-length location-1 location-5) 1)
  (road location-1 location-6)
  (= (road-length location-1 location-6) 1)
  (road location-1 location-7)
  (= (road-length location-1 location-7) 1)
  (road location-1 location-8)
  (= (road-length location-1 location-8) 1)
  (road location-1 location-9)
  (= (road-length location-1 location-9) 1)
  (road location-1 location-10)
  (= (road-length location-1 location-10) 1)
  (road location-2 location-1)
  (= (road-length location-2 location-1) 1)
  (road location-2 location-3)
  (= (road-length location-2 location-3) 1)
  (road location-2 location-4)
  (= (road-length location-2 location-4) 1)
  (road location-2 location-5)
  (= (road-length location-2 location-5) 1)
  (road location-2 location-6)
  (= (road-length location-2 location-6) 1)
  (road location-2 location-7)
  (= (road-length location-2 location-7) 1)
  (road location-2 location-8)
  (= (road-length location-2 location-8) 1)
  (road location-2 location-9)
  (= (road-length location-2 location-9) 1)
  (road location-2 location-10)
  (= (road-length location-2 location-10) 1)
  (road location-3 location-1)
  (= (road-length location-3 location-1) 1)
  (road location-3 location-2)
  (= (road-length location-3 location-2) 1)
  (road location-3 location-4)
  (= (road-length location-3 location-4) 1)
  (road location-3 location-5)
  (= (road-length location-3 location-5) 1)
  (road location-3 location-6)
  (= (road-length location-3 location-6) 1)
  (road location-3 location-7)
  (= (road-length location-3 location-7) 1)
  (road location-3 location-8)
  (= (road-length location-3 location-8) 1)
  (road location-3 location-9)
  (= (road-length location-3 location-9) 1)
  (road location-3 location-10)
  (= (road-length location-3 location-10) 1)
  (road location-4 location-1)
  (= (road-length location-4 location-1) 1)
  (road location-4 location-2)
  (= (road-length location-4 location-2) 1)
  (road location-4 location-3)
  (= (road-length location-4 location-3) 1)
  (road location-4 location-5)
  (= (road-length location-4 location-5) 1)
  (road location-4 location-6)
  (= (road-length location-4 location-6) 1)
  (road location-4 location-7)
  (= (road-length location-4 location-7) 1)
  (road location-4 location-8)
  (= (road-length location-4 location-8) 1)
  (road location-4 location-9)
  (= (road-length location-4 location-9) 1)
  (road location-4 location-10)
  (= (road-length location-4 location-10) 1)
  (road location-5 location-1)
  (= (road-length location-5 location-1) 1)
  (road location-5 location-2)
  (= (road-length location-5 location-2) 1)
  (road location-5 location-3)
  (= (road-length location-5 location-3) 1)
  (road location-5 location-4)
  (= (road-length location-5 location-4) 1)
  (road location-5 location-6)
  (= (road-length location-5 location-6) 1)
  (road location-5 location-7)
  (= (road-length location-5 location-7) 1)
  (road location-5 location-8)
  (= (road-length location-5 location-8) 1)
  (road location-5 location-9)
  (= (road-length location-5 location-9) 1)
  (road location-5 location-10)
  (= (road-length location-5 location-10) 1)
  (road location-6 location-1)
  (= (road-length location-6 location-1) 1)
  (road location-6 location-2)
  (= (road-length location-6 location-2) 1)
  (road location-6 location-3)
  (= (road-length location-6 location-3) 1)
  (road location-6 location-4)
  (= (road-length location-6 location-4) 1)
  (road location-6 location-5)
  (= (road-length location-6 location-5) 1)
  (road location-6 location-7)
  (= (road-length location-6 location-7) 1)
  (road location-6 location-8)
  (= (road-length location-6 location-8) 1)
  (road location-6 location-9)
  (= (road-length location-6 location-9) 1)
  (road location-6 location-10)
  (= (road-length location-6 location-10) 1)
  (road location-7 location-1)
  (= (road-length location-7 location-1) 1)
  (road location-7 location-2)
  (= (road-length location-7 location-2) 1)
  (road location-7 location-3)
  (= (road-length location-7 location-3) 1)
  (road location-7 location-4)
  (= (road-length location-7 location-4) 1)
  (road location-7 location-5)
  (= (road-length location-7 location-5) 1)
  (road location-7 location-6)
  (= (road-length location-7 location-6) 1)
  (road location-7 location-8)
  (= (road-length location-7 location-8) 1)
  (road location-7 location-9)
  (= (road-length location-7 location-9) 1)
  (road location-7 location-10)
  (= (road-length location-7 location-10) 1)
  (road location-8 location-1)
  (= (road-length location-8 location-1) 1)
  (road location-8 location-2)
  (= (road-length location-8 location-2) 1)
  (road location-8 location-3)
  (= (road-length location-8 location-3) 1)
  (road location-8 location-4)
  (= (road-length location-8 location-4) 1)
  (road location-8 location-5)
  (= (road-length location-8 location-5) 1)
  (road location-8 location-6)
  (= (road-length location-8 location-6) 1)
  (road location-8 location-7)
  (= (road-length location-8 location-7) 1)
  (road location-8 location-9)
  (= (road-length location-8 location-9) 1)
  (road location-8 location-10)
  (= (road-length location-8 location-10) 1)
  (road location-9 location-1)
  (= (road-length location-9 location-1) 1)
  (road location-9 location-2)
  (= (road-length location-9 location-2) 1)
  (road location-9 location-3)
  (= (road-length location-9 location-3) 1)
  (road location-9 location-4)
  (= (road-length location-9 location-4) 1)
  (road location-9 location-5)
  (= (road-length location-9 location-5) 1)
  (road location-9 location-6)
  (= (road-length location-9 location-6) 1)
  (road location-9 location-7)
  (= (road-length location-9 location-7) 1)
  (road location-9 location-8)
  (= (road-length location-9 location-8) 1)
  (road location-9 location-10)
  (= (road-length location-9 location-10) 1)
  (road location-10 location-1)
  (= (road-length location-10 location-1) 1)
  (road location-10 location-2)
  (= (road-length location-10 location-2) 1)
  (road location-10 location-3)
  (= (road-length location-10 location-3) 1)
  (road location-10 location-4)
  (= (road-length location-10 location-4) 1)
  (road location-10 location-5)
  (= (road-length location-10 location-5) 1)
  (road location-10 location-6)
  (= (road-length location-10 location-6) 1)
  (road location-10 location-7)
  (= (road-length location-10 location-7) 1)
  (road location-10 location-8)
  (= (road-length location-10 location-8) 1)
  (road location-10 location-9)
  (= (road-length location-10 location-9) 1)
  (at truck-1 location-1)
  (capacity truck-1 capacity-4)
  (at truck-2 location-2)
  (capacity truck-2 capacity-4)
  (at truck-3 location-3)
  (capacity truck-3 capacity-4)
  (at package-1 location-1)
  (at package-2 location-1)
  (at package-3 location-1)
  (at package-4 location-1)
  (at package-5 location-2)
  (at package-6 location-2)
  (at package-7 location-2)
  (at package-8 location-3)
  (at package-9 location-3)
  (at package-10 location-3)
 )
 (:goal
  (and
    (at package-1 location-1)
    (at package-2 location-4)
    (at package-3 location-8)
    (at package-4 location-5)
    (at package-5 location-3)
    (at package-6 location-2)
    (at package-7 location-6)
    (at package-8 location-9)
    (at package-9 location-7)
    (at package-10 location-10)
  )
 )
 (:metric minimize (total-cost))
)
\end{lstlisting}

\begin{lstlisting}[language=lisp, caption={Example Domain Description for Transport}, label={lst:transport_dd}]
    I am moving packages between locations using vehicles. Here are the actions I can do

Drive a vehicle from one location to another
Pick up a package with a vehicle
Drop a package from a vehicle at a location

I have the following restrictions on my actions:
I can only drive a vehicle from one location to another if there is a road connecting the two locations.
I can only pick up a package if the vehicle and the package are at the same location.
I can only pick up a package if the vehicle has enough remaining carrying capacity.
Once I pick up a package, the package is inside the vehicle and is no longer at the location.
I can only drop a package if the package is currently inside the vehicle.
I can only drop a package at the location where the vehicle currently is.
Once I drop a package, the package is at that location and is no longer inside the vehicle.
Driving a vehicle has a cost that depends on the road length.
Picking up and dropping a package each have a cost of 1.
\end{lstlisting}

\begin{lstlisting}[language=lisp, caption={Example Problem Description for Transport}, label={lst:transport_pd}]
    As initial conditions I have 10 locations numbered 1 to 10, 10 packages numbered 1 to 10, and 3 vehicles numbered 1 to 3. The locations form one fully connected road network, so every location is directly connected to every other location in both directions. Vehicle 1 starts at location 1, vehicle 2 starts at location 2, and vehicle 3 starts at location 3. Packages 1 through 4 are initially at location 1, packages 5 through 7 are initially at location 2, and packages 8 through 10 are initially at location 3. All 3 vehicles have carrying capacity 4, and every road has length 1.
My goal is to have that package 1 is at location 1, package 2 is at location 4, package 3 is at location 8, package 4 is at location 5, package 5 is at location 3, package 6 is at location 2, package 7 is at location 6, package 8 is at location 9, package 9 is at location 7, and package 10 is at location 10.
\end{lstlisting}

\begin{lstlisting}[language=lisp, caption={Prompt for LLM-as-Formalizer for BlocksWorld}, label={lst:blocksworld_formalizer_prompt}]
PDDL problem file contains problem name, domain name, objects in this problem instance, init state of objects, and goal state of objects.
Based on the natural language problem description, identify the relevant objects for this problem with their names and types.
Represent the initial state with the appropriate predicates and object arguments. Represent the goal state with the appropriate predicates and object arguments.
PDDL problem file has a definitive syntax that must be followed for any problem. An abstract example PDDL problem file is given below.

<problem_file>
(define
	(problem problem_name)
	(:domain domain_name)
	(:objects
		obj1 obj2 - type1
		obj3, obj4 - type2
	)
	(:init (predicate1 obj1 obj3) (predicate2 obj2 obj3))
	(:goal (and (predicate1 obj1 obj4) (predicate2 obj2 obj3)))
)
</problem_file>

Notes for generating problem file:
- obj1, obj2, ... are only representative and should be replaced with appropriate objects. There could be any number of objects with their types.
- init state with predicate1 & predicate2 is only representative and should be replaced with appropriate predicates that define init state
- goal state with predicate1 & predicate2 is only representative and should be replaced with appropriate predicates that define goal state
- predicates with proper arguments could be combined to combine complex boolean expression to represent init and goal states 
- The braces should be balanced for each section of the PDDL program
- Use predicates with arguments of the right type as declared in domain file
- All the objects that would be arguments of predicates in init and goal states should be declared in :objects


Additional strict constraints for large instances (55-100 blocks):
- Output only one complete problem PDDL wrapped in <problem_file>...</problem_file>. Do not output any domain file.
- Include every block mentioned in the problem description exactly once in :objects, and do not include undeclared objects.
- Use only predicates and action-relevant symbols that are valid for the provided domain PDDL.
- In :init, ensure each block appears in exactly one support relation: either (on-table blockX) or (on blockX blockY), but not both.
- In :init, include (clear blockX) if and only if blockX has no block on top of it.
- In :init, include exactly one (arm-empty).
- Ensure there are no self-relations such as (on blockX blockX).
- In :goal, use only declared objects and valid predicates, and avoid contradictory goals.
- Ensure parentheses are balanced and PDDL syntax is valid.

;; Example Second Stage Prompt
Review the first draft of the BlocksWorld problem file against the original task and the fixed BlocksWorld rules, then produce the corrected final problem file.

A draft can preserve much of a large arrangement while quietly changing a few stack bases, top blocks, or requested final relationships. Pay particular attention to the following:
- Preserve every block name exactly as given and account for exactly the blocks described by the task, with none missing and none added.
- Reproduce the starting arrangement stack by stack. For each stack, the lowest block rests directly on the table, every higher block rests on the correct block immediately below it, and a block resting on another block must not also be treated as a stack base.
- Identify as unobstructed only the true top block of each starting stack. No block with another block above it should be described as unobstructed.
- Do not copy details from the requested final arrangement into the starting arrangement unless the task explicitly says they are already true at the start.
- Preserve the requested final arrangement exactly, including every stated direct support relationship and every block that must rest directly on the table. Do not omit a final stack base merely because most of the description concerns blocks resting on other blocks.
- For long stacks, verify every neighboring pair and the bottom block directly against the task instead of inventing a shorter numerical, alternating, or repeating pattern.
- Keep the starting arrangement and requested final arrangement independent. Their stack bases, top blocks, and support order may differ.
- Do not introduce any relationship or condition that is not supported by the fixed rules or the original task. Preserve every part of the draft that already agrees with them.

Return the complete corrected problem file in one <problem_file>...</problem_file> block, with no text before or after it.

Original task prompt:
{original_prompt}
\end{lstlisting}

\begin{lstlisting}[language=lisp, caption={Prompt for LLM-as-Planner for BlocksWorld}, label={lst:blocksworld_planner_prompt}]
You are an expert automated planner. Your task is to read a natural-language domain description, the corresponding problem description, and the provided ground-truth domain PDDL, reason through the objects, actions, and goals, and output a valid plan.

Recall the distinctions:
- The domain description explains the general world: available action schemas, predicates, and object types.
- The problem description gives the concrete instance: specific objects, the initial state, and the goals for this planning episode.
- The ground-truth domain PDDL provides the formal acton schema and predicate definitions that your plan must follow exactly.

Planning output requirements:
- Produce the final plan inside <plan>...</plan>. Each action must be on its own line in classical PDDL plan syntax, e.g. ,(move robot1 roomA roomB).
- Maintain the execution order from top to bottom. Do not include step numbers, timestamps, or probabilities.
- Only emit actions that are applicable given the initial state and that eventually achieve the goals.
- After the closing </plan> tag, do not add extra commentary.

Example format (replace names with the task-specific details):
<plan>
(pick-up block1 hand)
(stack block1 block2)
(move robot room1 room2)
</plan>

Follow this template for every problem. EOF
\end{lstlisting}

\begin{lstlisting}[language=lisp, caption={Prompt for D\&C Problem File Header}, label={lst:blocksworld_dnc_header_prompt}]
PDDL problem file's header parts contain problem name, domain name, and objects in this problem instance.
Based on the natural language problem description, identify the relevant objects for this problem with their names and types..
PDDL problem file header has a definitive syntax that must be followed for any problem. An abstract example PDDL problem file header is given below.

<header>
(define
	(problem problem_name)
	(:domain domain_name)
	(:objects
		obj1 obj2 - type1
		obj3, obj4 - type2
	)
)
</header>

Notes for generating problem file header:
- obj1, obj2, ... are only representative and should be replaced with appropriate objects. There could be any number of objects with their types. 
- The braces should be balanced for each section of the PDDL program header
- All the objects should be declared in :objects


Additional strict constraints for all instances (5-100 blocks):
- Output only one complete problem PDDL header wrapped in <header>...</header>. Do not output any domain file.
- Include every block mentioned in the problem description once in :objects, and do not include undeclared objects.
- Only write for the header parts.
- Do not write anything for :init.
- Do not write anything for :goal.
- Ensure parentheses are balanced and PDDL syntax is valid.
\end{lstlisting}

\begin{lstlisting}[language=lisp, caption={Prompt for D\&C Problem File Main Body}, label={lst:blocksworld_dnc_body_prompt}]
PDDL problem file fact states contains init state of objects, and goal state of objects.
Based on the natural language problem description, identify the relevant init and goal states for this problem ONE-BY-ONE.
Represent the initial state with the appropriate predicates and object arguments. Represent the goal state with the appropriate predicates and object arguments.
PDDL problem file fact state has a definitive syntax that must be followed for any problem. An few abstract examples PDDL problem file fact state is given below.

<fact>
(predicate1 obj1 obj3) (predicate2 obj2 obj3))
</fact>
for init state

OR

<fact>
(and (predicate1 obj1 obj4) (predicate2 obj2 obj3)))
</fact>
for goal state


Notes for generating problem file fact state:
- obj1, obj2, ... are only representative and should be replaced with appropriate objects. There could be any number of objects with their types.
- init state with predicate1 & predicate2 is only representative and should be replaced with appropriate predicates that define init state.
- goal state with predicate1 & predicate2 is only representative and should be replaced with appropriate predicates that define goal state.
- predicates with proper arguments could be combined to combine complex boolean expression to represent init and goal states. 
- The braces should be balanced for each fact state of the PDDL program.
- Use predicates with arguments of the right type as declared in domain file.


Additional strict constraints for ALL instances (5-100 blocks):
- Output only one problem PDDL fact state wrapped in <fact>...</fact>. Do not output any domain file.
- DO NOT WRITE ANY :init or :goal inside <fact>...</fact>!!!
- Use only predicates and action-relevant symbols that are valid for the provided domain PDDL.
- Section headers :init and :goal are only produced once.
- For init, ensure the one fact is a correct translation of the corresponding fact in the natural language problem description.
- For init, write (clear blockX) if and only if blockX is clear. 
- For init, write (arm-empty) if and only if the hand is empty.
- If a fact belongs to section :init, output one init fact atom.
- Ensure there are no self-relations such as (on blockX blockX).
- For goal, use only declared objects and valid predicates, and avoid contradictory goals.
- If a fact belongs to section :goal, output one goal fact atom.
- Ensure parentheses are balanced and PDDL syntax is valid.

Important examples for incorrect and correct atomic fact:
- INVALID: (:init (clear blockX))
- INVALID: (:goal (on-table blockX))
- Valid: (clear blockX)
- Valid: (on-table blockX)
\end{lstlisting}

\begin{lstlisting}[language=lisp, caption={Prompt for LLM-as-Higher-Order-Formalizer for BlocksWorld}, label={lst:blocksworld_higher_order_formalizer_prompt}]
;; Original prompt
PDDL problem file contains problem name, domain name, objects in this problem instance, init state of objects, and goal state of objects.
Based on the natural language problem description, identify the relevant objects for this problem with their names and types.
In a formal PDDL problem file, initial state have the appropriate predicates and object arguments.
In a formal PDDL problem file, goal state also have the appropriate predicates and object arguments.
PDDL problem file has a definitive syntax that must be followed for any problem. An abstract example PDDL problem file is given below.

(define
	(problem problem_name)
	(:domain domain_name)
	(:objects
		obj1 obj2 - type1
		obj3, obj4 - type2
	)
	(:init (predicate1 obj1 obj3) (predicate2 obj2 obj3))
	(:goal (and (predicate1 obj1 obj4) (predicate2 obj2 obj3)))
)

You MUST generate a PYTHON SCRIPT that, upon execution, generates the PDDL problem file that matches the natural language problem description:
<generator>
the python script code...
</generator>

EXAMPLE executable generator style (generalized):

<generator>
def build_init_from_unraveling_rule(n):
  blocks = [f"block{i}" for i in range(1, n + 1)]

  # Odd and even stacks are TOP -> BOTTOM with increasing numbering
  odd_stack = [b for i, b in enumerate(blocks, start=1) if i % 2 == 1]
  even_stack = [b for i, b in enumerate(blocks, start=1) if i % 2 == 0]

  init_facts = []

  # Convert one stack (top->bottom) to on/on-table/clear facts
  def emit_stack(stack):
      if not stack:
          return
      for idx, block in enumerate(stack):
          if idx == len(stack) - 1:
              init_facts.append(f"(on-table {block})")
          else:
              under = stack[idx + 1]
              init_facts.append(f"(on {block} {under})")
          if idx == 0:
              init_facts.append(f"(clear {block})")

  emit_stack(odd_stack)
  emit_stack(even_stack)

  init_facts.append("(arm-empty)")
  return blocks, init_facts


def build_problem_pddl(problem_name, domain_name, blocks, init_facts, goal_facts):
  lines = []
  lines.append(f"(define (problem {problem_name})")
  lines.append(f"  (:domain {domain_name})")
  lines.append(f"  (:objects {' '.join(blocks)})")
  lines.append("  (:init")
  for fact in init_facts:
      lines.append(f"    {fact}")
  lines.append("  )")
  lines.append("  (:goal (and")
  for fact in goal_facts:
      lines.append(f"    {fact}")
  lines.append("  ))")
  lines.append(")")
  return "\n".join(lines) + "\n"


def main():
  # n should be inferred from the given NL problem description in the real task
  n = 10
  problem_name = "example_problem"
  domain_name = "blocksworld"

  blocks, init_facts = build_init_from_unraveling_rule(n)

  # Goal extraction can be abstracted; replace with parsed goal facts from the given NL problem
  goal_facts = [
      "(on block1 block5)",
      "(on block2 block1)",
      "(on-table block4)"
  ]

  pddl_text = build_problem_pddl(problem_name, domain_name, blocks, init_facts, goal_facts)

  with open("problem.pddl", "w", encoding="utf-8") as f:
      f.write(pddl_text)


if __name__ == "__main__":
  main()
</generator>


Notes for generating the python script:
- The Python script is ONLY TRANSLATING the natural language problem description to the formal PDDL problem file.
- The Python script is ABSOLUTELY NOT SOLVING the BlocksWorld problem


Additional strict constraints for all instances (5-100 blocks):
- Output only one complete Python script wrapped in <generator>...</generator>. Do not output any domain file.
- The script MUST derive init/goal facts algorithmically from compact structure, not hardcoded full fact lists.
- Disallowed: manually listing full init_state/goal_state facts as constants.
- Required: at least one loop that constructs predicates (e.g., on/on-table/clear) from computed stacks or parsed patterns.
- Required: compute object list from inferred block count (do not hardcode all object names manually).
- The Python script, upon execution, need to produce a complete PDDL problem file that includes every block mentioned in the problem description once in :objects and do not include undeclared objects.
- The Python script, upon execution, need to produce a complete PDDL problem file that uses only predicates and action-relevant symbols that are valid for the provided domain PDDL.
- The Python script, upon execution, need to produce a complete PDDL problem file that in :init, ensures each block appears in one support relation: either (on-table blockX) or (on blockX blockY), but not both.
- The Python script, upon execution, need to produce a complete PDDL problem file that in :init, include (clear blockX) if and only if blockX has no block on top of it.
- The Python script, upon execution, need to produce a complete PDDL problem file that in :init, include one (arm-empty).
- The Python script, upon execution, need to produce a complete PDDL problem file that ensures there are no self-relations such as (on blockX blockX).
- The Python script, upon execution, need to produce a complete PDDL problem file that in :goal, use only declared objects and valid predicates, and avoid contradictory goals.
- The Python script, upon execution, need to produce a complete PDDL problem file that ensures parentheses are balanced and PDDL syntax is valid.

;; Example Second Stage Prompt
Review the existing first-stage Python generator and the problem file it produced against the original task prompt and fixed domain. Return a corrected, complete Python generator. 

Pay special attention to the Python logic that constructs identifiers and the desired arrangement. Preserve the complete textual form required for every block identifier, never replace an identifier with a bare number or remove its shared word-based prefix. Generate exactly the conditions explicitly requested for the desired arrangement, without inferring additional conditions about which blocks have nothing above them or whether the robot hand is empty unless the task states them. When the script determines which blocks rest directly on the surface, remember that a lower block may rest on the surface while supporting another block, exclude only blocks that are placed on another block, not every block participating in a stacking relationship. Inspect set comprehensions, exclusion sets, and loops carefully so that they neither omit requested conditions nor invent new ones. Correct the generator itself so that these requirements are reflected throughout the problem file it produces, while preserving parts of the draft that already agree with the task and domain.

Return the complete corrected Python generator in one <generator>...</generator> block, with no text before or after it. The generator must write the final problem file to problem.pddl in the current working directory.

Original task prompt:
{original_prompt}

Existing first-stage generator.py:
{draft_generator_code}

Problem file produced by the existing generator:
{draft_problem_pddl}
\end{lstlisting}

\begin{lstlisting}[language=lisp, caption={Prompt for LLM-as-Planner for ChildSnack}, label={lst:prompt_planner_childsnack}]
    You are an expert automated planner. Your task is to read a natural-language domain description, the corresponding problem description, and the provided ground-truth domain PDDL, reason through the objects, actions, and goals, and output a valid plan.

Recall the distinctions:
- The domain description explains the general world: available action schemas, predicates, and object types.
- The problem description gives the concrete instance: specific objects, the initial state, and the goals for this planning episode.
- The ground-truth domain PDDL provides the formal acton schema and predicate definitions that your plan must follow exactly.

Planning output requirements:
- Produce the final plan inside <plan>...</plan>. Each action must be on its own line in classical PDDL plan syntax, e.g. `(move robot1 roomA roomB)`.
- Maintain the execution order from top to bottom. Do not include step numbers, timestamps, or probabilities.
- Only emit actions that are applicable given the initial state and that eventually achieve the goals.
- After the closing </plan> tag, do not add extra commentary.

Example format (replace names with the task-specific details):
<plan>
(make_sandwich_no_gluten sandw1 bread1 content1)
(put_on_tray sandw1 tray1)
(move_tray tray1 kitchen table1)
(serve_sandwich_no_gluten sandw1 child1 tray1 table1)
</plan>

Follow this template for every problem. EOF
\end{lstlisting}

\begin{lstlisting}[language=lisp, caption={Prompt for LLM-as-Formalizer for ChildSnack}, label={lst:prompt_formalizer_childsnack}]
    PDDL problem file contains problem name, domain name, objects in this problem instance, init state of objects, and goal state of objects.
Based on the natural language problem description, identify the relevant objects for this problem with their names and types.
Represent the initial state with the appropriate predicates and object arguments. Represent the goal state with the appropriate predicates and object arguments.
PDDL problem file has a definitive syntax that must be followed for any problem. An abstract example PDDL problem file is given below.

<problem_file>
(define
	(problem problem_name)
	(:domain domain_name)
	(:objects
		obj1 obj2 - type1
		obj3, obj4 - type2
	)
	(:init (predicate1 obj1 obj3) (predicate2 obj2 obj3))
	(:goal (and (predicate1 obj1 obj4) (predicate2 obj2 obj3)))
)
</problem_file>

Notes for generating problem file:
- obj1, obj2, ... are only representative and should be replaced with appropriate objects. There could be any number of objects with their types.
- init state with predicate1 & predicate2 is only representative and should be replaced with appropriate predicates that define init state
- goal state with predicate1 & predicate2 is only representative and should be replaced with appropriate predicates that define goal state
- predicates with proper arguments could be combined to combine complex boolean expression to represent init and goal states 
- The braces should be balanced for each section of the PDDL program
- Use predicates with arguments of the right type as declared in domain file
- All the objects that would be arguments of predicates in init and goal states should be declared in :objects


Additional strict constraints for large instances (5-30 children):
- Output only one complete problem PDDL wrapped in <problem_file>...</problem_file>. Do not output any domain file.
- Include every required problem object once in :objects, and do not include undeclared or duplicate objects.
- Use only predicates and action-relevant symbols that are valid for the provided domain PDDL.
- In this ChildSnack formulation, declare all required children, bread-portion objects, content-portion objects, sandwich objects, tray objects, and place objects once using valid names and valid types from the problem description.
- Do not declare the kitchen as a problem object if it is already provided as a domain constant, use it only where valid predicates require it.
- In :init, include the required at facts for trays, the required at_kitchen_bread facts for bread portions, the required at_kitchen_content facts for content portions, and the required notexist facts for sandwich objects using only declared objects and valid predicates.
- In :init, assign each child one valid gluten-status predicate, and include the required no_gluten_bread and no_gluten_content facts for the intended gluten-free ingredients.
- In :init, include one valid waiting fact for each child at the intended place, using only declared objects and valid predicates.
- In :goal, include the required served facts for all and only the intended children, using only declared objects and valid predicates, and avoid contradictory goals.
- Ensure parentheses are balanced and PDDL syntax is valid.

;; Example Second Stage Prompt
Review the first draft of the ChildSnack problem file against the original task prompt and the fixed rules, then produce the corrected final problem file.

A draft can appear broadly plausible while still changing small but important details. Pay particular attention to the following:
- Preserve every name exactly as it appears in the original task, and keep each child, bread portion, filling portion, sandwich, tray, and place in its stated role. Do not replace short names with longer or more natural-sounding alternatives.
- Use each sandwich name consistently whenever that sandwich is mentioned, including when describing which sandwiches have not yet been prepared.
- Account for exactly the bread portions and filling portions that the task says begin in the kitchen, without adding or omitting any.
- Preserve every tray and the precise place where it begins.
- Follow the exact gluten-free status of the bread and filling portions and the exact allergy status of the children. Pay close attention to where each described pattern begins and to any stated exclusions. Do not continue a repeating pattern from an assumed starting point.
- Preserve each child's stated waiting place. Do not reorganize the children into increasing groups, reverse groups, or any other intuitive arrangement unless the original task explicitly gives that arrangement.
- Ensure that exactly the children identified by the task are expected to be served, with nobody missing and nobody added.
- Do not introduce any kind of relationship or condition that is not supported by the fixed rules. Preserve every part of the draft that already agrees with the original task.

Return the complete corrected problem file in one <problem_file>...</problem_file> block, with no text before or after it.

Original task prompt:
{original_prompt}
\end{lstlisting}

\begin{lstlisting}[language=lisp, caption={Prompt for LLM-as-Higher-Order-Formalizer for ChildSnack}, label={lst:prompt_ho_formalizer_childsnack}]
;; Original Prompt
PDDL problem file contains problem name, domain name, objects in this problem instance, init state of objects, and goal state of objects.
Based on the natural language problem description, identify the relevant objects for this problem with their names and types.
In a formal PDDL problem file, initial state have the appropriate predicates and object arguments.
In a formal PDDL problem file, goal state also have the appropriate predicates and object arguments.
PDDL problem file has a definitive syntax that must be followed for any problem. An abstract example PDDL problem file is given below.

(define
	(problem problem_name)
	(:domain domain_name)
	(:objects
		obj1 obj2 - type1
		obj3, obj4 - type2
	)
	(:init (predicate1 obj1 obj3) (predicate2 obj2 obj3))
	(:goal (and (predicate1 obj1 obj4) (predicate2 obj2 obj3)))
)

You MUST generate a PYTHON SCRIPT that, upon execution, generates the PDDL problem file that matches the natural language problem description:
<generator>
the python script code...
</generator>

EXAMPLE executable generator style from another domain (generalized):

<generator>
def build_init_from_unraveling_rule(n):
  blocks = [f"block{i}" for i in range(1, n + 1)]

  # Odd and even stacks are TOP -> BOTTOM with increasing numbering
  odd_stack = [b for i, b in enumerate(blocks, start=1) if i % 2 == 1]
  even_stack = [b for i, b in enumerate(blocks, start=1) if i % 2 == 0]

  init_facts = []

  # Convert one stack (top->bottom) to on/on-table/clear facts
  def emit_stack(stack):
      if not stack:
          return
      for idx, block in enumerate(stack):
          if idx == len(stack) - 1:
              init_facts.append(f"(on-table {block})")
          else:
              under = stack[idx + 1]
              init_facts.append(f"(on {block} {under})")
          if idx == 0:
              init_facts.append(f"(clear {block})")

  emit_stack(odd_stack)
  emit_stack(even_stack)

  init_facts.append("(arm-empty)")
  return blocks, init_facts


def build_problem_pddl(problem_name, domain_name, blocks, init_facts, goal_facts):
  lines = []
  lines.append(f"(define (problem {problem_name})")
  lines.append(f"  (:domain {domain_name})")
  lines.append(f"  (:objects {' '.join(blocks)})")
  lines.append("  (:init")
  for fact in init_facts:
      lines.append(f"    {fact}")
  lines.append("  )")
  lines.append("  (:goal (and")
  for fact in goal_facts:
      lines.append(f"    {fact}")
  lines.append("  ))")
  lines.append(")")
  return "\n".join(lines) + "\n"


def main():
  # n should be inferred from the given NL problem description in the real task
  n = 10
  problem_name = "example_problem"
  domain_name = "xxx"

  blocks, init_facts = build_init_from_unraveling_rule(n)

  # Goal extraction can be abstracted; replace with parsed goal facts from the given NL problem
  goal_facts = [
      "(on block1 block5)",
      "(on block2 block1)",
      "(on-table block4)"
  ]

  pddl_text = build_problem_pddl(problem_name, domain_name, blocks, init_facts, goal_facts)

  with open("problem.pddl", "w", encoding="utf-8") as f:
      f.write(pddl_text)


if __name__ == "__main__":
  main()
</generator>


Notes for generating the python script:
- The Python script is ONLY TRANSLATING the natural language problem description to the formal PDDL problem file.
- The Python script is ABSOLUTELY NOT SOLVING the ChildSnack problem


Additional strict constraints for all instances (5-30 children):
- Output only one complete Python script wrapped in <generator>...</generator>. Do not output any domain file.
- The script MUST derive init/goal facts algorithmically from compact structure, not hardcoded full fact lists.
- Disallowed: manually listing full init_state/goal_state facts as constants.
- Required: at least one loop that constructs predicates (e.g., waiting/at_kitchen_bread/served) from parsed patterns.
- Required: compute object list from inferred child count and the fixed template counts for breads, contents, sandwiches, trays, and tables (do not hardcode all object names manually).
- The Python script, upon execution, need to produce a complete PDDL problem file that includes every required child, bread-portion, content-portion, tray, table, and sandwich object once in :objects and does not include undeclared problem objects.
- The Python script, upon execution, need to produce a complete PDDL problem file that uses only predicates and object arguments that are valid for the provided ChildSnack domain PDDL.
- The Python script, upon execution, need to produce a complete PDDL problem file that in :init, places every tray at one valid place, with the kitchen referenced only as the domain constant kitchen and not redeclared in :objects.
- The Python script, upon execution, need to produce a complete PDDL problem file that in :init, places every bread portion once with at_kitchen_bread and every content portion once with at_kitchen_content.
- The Python script, upon execution, need to produce a complete PDDL problem file that in :init, marks each sandwich placeholder with notexist once and does not also place that sandwich at the kitchen or on a tray initially.
- The Python script, upon execution, need to produce a complete PDDL problem file that in :init, assigns each child one gluten-status fact, either allergic_gluten or not_allergic_gluten, but not both.
- The Python script, upon execution, need to produce a complete PDDL problem file that in :init, assigns no gluten bread and no gluten content to the portions required by the described template pattern, without adding those facts to regular portions.
- The Python script, upon execution, need to produce a complete PDDL problem file that in :init, places every child in one waiting fact at one valid table according to the described tray-assignment pattern.
- The Python script, upon execution, need to produce a complete PDDL problem file that in :goal, includes the served facts for all described children and avoids contradictory or undeclared goals.
- The Python script, upon execution, need to produce a complete PDDL problem file that ensures parentheses are balanced and PDDL syntax is valid.

;; Example Second Stage Prompt 
Review the existing first-stage Python generator and the problem file it produced against the original task prompt and fixed domain. Return a corrected, complete Python generator.

Pay special attention to recurring groups described as beginning with a particular member. Treat that named member as part of the group and continue with the stated spacing, without shifting the group to a later member. Keep each dietary requirement consistent with the corresponding bread and content portions identified in the task. Preserve the entity names established by the original description rather than shortening or replacing them. Check that those names remain consistent among the declared entities, starting situation, and desired result. Correct the generator itself so that any repaired interpretation is reflected throughout the problem file it produces, while preserving parts of the draft that already agree with the task and domain.

Return the complete corrected Python generator in one <generator>...</generator> block, with no text before or after it. The generator must write the final problem file to problem.pddl in the current working directory.

Original task prompt:
{original_prompt}

Existing first-stage generator.py:
{draft_generator_code}

Problem file produced by the existing generator:
{draft_problem_pddl}
\end{lstlisting}

\begin{lstlisting}[language=lisp, caption={Prompt for LLM-as-Planner for Elevators}, label={lst:prompt_planner_elevators}]
    You are an expert automated planner. Your task is to read a natural-language domain description, the corresponding problem description, and the provided ground-truth domain PDDL, reason through the objects, actions, and goals, and output a valid plan.

Recall the distinctions:
- The domain description explains the general world: available action schemas, predicates, and object types.
- The problem description gives the concrete instance: specific objects, the initial state, and the goals for this planning episode.
- The ground-truth domain PDDL provides the formal acton schema and predicate definitions that your plan must follow exactly.

Planning output requirements:
- Produce the final plan inside <plan>...</plan>. Each action must be on its own line in classical PDDL plan syntax, e.g. (move robot1 roomA roomB).
- Maintain the execution order from top to bottom. Do not include step numbers, timestamps, or probabilities.
- Only emit actions that are applicable given the initial state and that eventually achieve the goals.
- After the closing </plan> tag, do not add extra commentary.

Example format (replace names with the task-specific details):
<plan>
(board p1 slow1 n0 n0 n1)
(move-up-slow slow1 n0 n9)
(leave p1 slow1 n9 n1 n0)
</plan>

Follow this template for every problem. EOF
\end{lstlisting}

\begin{lstlisting}[language=lisp, caption={Prompt for LLM-as-Formalizer for Elevators}, label={lst:prompt_formalizer_elevators}]
    PDDL problem file contains problem name, domain name, objects in this problem instance, init state of objects, and goal state of objects.
Based on the natural language problem description, identify the relevant objects for this problem with their names and types.
Represent the initial state with the appropriate predicates and object arguments. Represent the goal state with the appropriate predicates and object arguments.
PDDL problem file has a definitive syntax that must be followed for any problem. An abstract example PDDL problem file is given below.

<problem_file>
(define
	(problem problem_name)
	(:domain domain_name)
	(:objects
		obj1 obj2 - type1
		obj3, obj4 - type2
	)
	(:init (predicate1 obj1 obj3) (predicate2 obj2 obj3))
	(:goal (and (predicate1 obj1 obj4) (predicate2 obj2 obj3)))
)
</problem_file>

Notes for generating problem file:
- obj1, obj2, ... are only representative and should be replaced with appropriate objects. There could be any number of objects with their types.
- init state with predicate1 & predicate2 is only representative and should be replaced with appropriate predicates that define init state
- goal state with predicate1 & predicate2 is only representative and should be replaced with appropriate predicates that define goal state
- predicates with proper arguments could be combined to combine complex boolean expression to represent init and goal states 
- The braces should be balanced for each section of the PDDL program
- Use predicates with arguments of the right type as declared in domain file
- All the objects that would be arguments of predicates in init and goal states should be declared in :objects


Additional strict constraints for large instances (10-100 floors):
- Output only one complete problem PDDL wrapped in <problem_file>...</problem_file>. Do not output any domain file.
- Include every required problem object once in :objects, and do not include undeclared or duplicate objects.
- Use only predicates and action-relevant symbols that are valid for the provided domain PDDL.
- In this Elevators formulation, declare all floor/count objects, passengers, slow elevators, and fast elevators once using valid names and valid types from the problem description.
- Floors numbered 1 through F in the natural-language description must be represented consistently by the domain's count objects n0 through n(F-1). These same count objects also represent elevator occupancy levels required by the domain.
- In :init, place all 15 passengers and all three elevators on floor 1, represent every elevator as initially carrying zero passengers, and do not initially mark any passenger as boarded.
- In :init, include every required can-hold fact: slow1 and slow2 can hold occupancy levels n1 through n4, and fast1 can hold occupancy levels n1 through n6.
- In :init, explicitly include every adjacent next relation between n0 and n(F-1), and every above relation from each lower floor object to every higher floor object. Do not omit relations, reverse their arguments, or include self relations.
- In :init, explicitly include a reachable-floor fact for every combination of declared elevator and declared floor.
- In :init, explicitly include one travel-slow and one travel-fast numeric assignment for every lower-floor/higher-floor pair. For floors separated by d levels, the slow cost must be d + 5 and the fast cost must be 3d + 1.
- Initialize total-cost to 0 and include (:metric minimize (total-cost)).
- In :goal, include one passenger-at destination fact for each of the 15 passengers, matching the destination floor stated for that passenger and using only declared objects.
- Ensure parentheses are balanced and PDDL syntax is valid.

;; Example Second Stage Prompt
Review the first draft of the elevator problem file against the original task and the fixed domain, then produce the corrected complete problem file.

Previous drafts have often looked plausible while containing missing required information, unsupported extra information, malformed numerical data, or incomplete large instances. Do not approve the draft merely because its overall shape looks reasonable. Before rewriting it, deliberately audit each of the following error categories.

Missing objects:
- Check that the object inventory contains every passenger, both slow elevators, the fast elevator, and every floor, count value required by the described building.
- Preserve every exact symbolic name and its stated category. The floor, count inventory must be one uninterrupted zero-based sequence from the bottom floor through the top floor.

Extra objects:
- Remove invented aliases, renamed elevators, generic replacements for the two elevator kinds, duplicate declarations, and objects not supported by the task.
- Do not split complete names into individual characters, repeat name prefixes, or silently substitute shorter names.

Missing initial-state information:
- Confirm that every passenger starts at the bottom, every elevator starts at the bottom, and every elevator initially carries zero passengers.
- Confirm that every elevator is associated with every floor, with no elevator or floor omitted.
- Confirm that each slow elevator allows every positive passenger count from one through four and that the fast elevator allows every positive passenger count from one through six. Include the complete range rather than only its maximum.
- Confirm that every neighboring pair across the entire floor sequence appears once in bottom-to-top order. The building sequence must not stop at an elevator capacity.
- Confirm that every distinct pair of floors appears once with the lower floor first and the higher floor second.
- For every such lower-to-higher pair, include both numerical travel costs. The slow-elevator cost is the number of floors traveled plus five, and the fast-elevator cost is three times that distance plus one.
- Confirm that the initial objective value is represented numerically and begins at zero.

Extra or incorrect initial-state information:
- Remove reversed copies, self-pairs, duplicate relationships, duplicate travel entries, and any relationship that places the higher floor before the lower floor.
- Do not include zero as a permitted passenger count, do not add counts beyond an elevator's capacity, and do not attach information to invented or wrongly named elevators.
- Do not rely on ordinary English intuition to reverse the required lower-floor-first ordering. Travel information is required only in that lower-to-higher direction.
- Treat every travel cost and the initial objective value as a numerical assignment associated with the intended quantity. Do not express a number as an additional ordinary argument, leave it detached, omit it, or provide an unvalued floor pair.

Missing, extra, or shifted destination requirements:
- Include one requested destination for every passenger described by the task, with no passenger missing, repeated, or added.
- Preserve each stated destination exactly. Convert a human-numbered floor to its zero-based internal floor once, do not leave it one-based and do not subtract an additional step from a value already converted.

Completeness and large-instance reliability:
- Include the requested optimization objective and every required part of the instance. Do not use placeholders, examples, summaries, ellipses, or comments claiming that omitted cases follow a pattern.
- Large buildings still require complete coverage of all elevators, all floors, all neighboring floor pairs, and all distinct lower-to-higher floor pairs with both costs. Use compact formatting if necessary, but never abbreviate or stop early.
- Ensure that the final answer contains one complete problem file and reaches its proper ending inside the required wrapper. Preserve everything already correct while repairing every missing, extra, malformed, reversed, mistyped, shifted, or truncated part.

Return only the complete corrected problem file in exactly one <problem_file>...</problem_file> block, with no text before or after it. Do not include Markdown fences or explanations.

Original task prompt:
{original_prompt}
\end{lstlisting}

\begin{lstlisting}[language=lisp, caption={Prompt for LLM-as-Higher-Order-Formalizer for Elevators}, label={lst:prompt_ho_formalizer_elevators}]
;; Original Prompt
PDDL problem file contains problem name, domain name, objects in this problem instance, init state of objects, and goal state of objects.
Based on the natural language problem description, identify the relevant objects for this problem with their names and types.
In a formal PDDL problem file, initial state have the appropriate predicates and object arguments.
In a formal PDDL problem file, goal state also have the appropriate predicates and object arguments.
PDDL problem file has a definitive syntax that must be followed for any problem. An abstract example PDDL problem file is given below.

(define
	(problem problem_name)
	(:domain domain_name)
	(:objects
		obj1 obj2 - type1
		obj3, obj4 - type2
	)
	(:init (predicate1 obj1 obj3) (predicate2 obj2 obj3))
	(:goal (and (predicate1 obj1 obj4) (predicate2 obj2 obj3)))
)

You MUST generate a PYTHON SCRIPT that, upon execution, generates the PDDL problem file that matches the natural language problem description:
<generator>
the python script code...
</generator>

EXAMPLE executable generator style from another domain (generalized):

<generator>
def build_init_from_unraveling_rule(n):
  blocks = [f"block{i}" for i in range(1, n + 1)]

  # Odd and even stacks are TOP -> BOTTOM with increasing numbering
  odd_stack = [b for i, b in enumerate(blocks, start=1) if i % 2 == 1]
  even_stack = [b for i, b in enumerate(blocks, start=1) if i % 2 == 0]

  init_facts = []

  # Convert one stack (top->bottom) to on/on-table/clear facts
  def emit_stack(stack):
      if not stack:
          return
      for idx, block in enumerate(stack):
          if idx == len(stack) - 1:
              init_facts.append(f"(on-table {block})")
          else:
              under = stack[idx + 1]
              init_facts.append(f"(on {block} {under})")
          if idx == 0:
              init_facts.append(f"(clear {block})")

  emit_stack(odd_stack)
  emit_stack(even_stack)

  init_facts.append("(arm-empty)")
  return blocks, init_facts


def build_problem_pddl(problem_name, domain_name, blocks, init_facts, goal_facts):
  lines = []
  lines.append(f"(define (problem {problem_name})")
  lines.append(f"  (:domain {domain_name})")
  lines.append(f"  (:objects {' '.join(blocks)})")
  lines.append("  (:init")
  for fact in init_facts:
      lines.append(f"    {fact}")
  lines.append("  )")
  lines.append("  (:goal (and")
  for fact in goal_facts:
      lines.append(f"    {fact}")
  lines.append("  ))")
  lines.append(")")
  return "\n".join(lines) + "\n"


def main():
  # n should be inferred from the given NL problem description in the real task
  n = 10
  problem_name = "example_problem"
  domain_name = "xxx"

  blocks, init_facts = build_init_from_unraveling_rule(n)

  # Goal extraction can be abstracted; replace with parsed goal facts from the given NL problem
  goal_facts = [
      "(on block1 block5)",
      "(on block2 block1)",
      "(on-table block4)"
  ]

  pddl_text = build_problem_pddl(problem_name, domain_name, blocks, init_facts, goal_facts)

  with open("problem.pddl", "w", encoding="utf-8") as f:
      f.write(pddl_text)


if __name__ == "__main__":
  main()
</generator>


Notes for generating the python script:
- The Python script is ONLY TRANSLATING the natural language problem description to the formal PDDL problem file.
- The Python script is ABSOLUTELY NOT SOLVING the Transport problem


Additional strict constraints for all instances (10-100 floors):
- Output only one complete Python script wrapped in <generator>...</generator>. Do not output any domain file.
- The script MUST derive init/goal facts algorithmically from compact structure, not hardcoded full fact lists.
- Disallowed: manually listing full init_state/goal_state facts as constants.
- Required: use loops to construct repeated predicates and numeric assignments, including floor relations, elevator reachability, passenger locations, and travel costs, from computed patterns.
- Required: compute object lists from the inferred floor count and the described passenger and elevator counts, do not hardcode complete object-name lists manually.
- The Python script, upon execution, need to produce a complete PDDL problem file that includes every floor/count object, passenger, slow elevator, and fast elevator required by the problem description once in :objects and does not use undeclared objects.
- The Python script, upon execution, need to produce a complete PDDL problem file that uses only predicates and action-relevant symbols that are valid for the provided domain PDDL.
- Floors numbered 1 through F in the natural-language description must be represented consistently by the domain's count objects n0 through n(F-1). These same count objects also represent elevator occupancy levels required by the domain.
- The Python script, upon execution, need to produce a complete PDDL problem file that in :init, places all 15 passengers and all three elevators on floor 1, represents every elevator as initially carrying zero passengers, and does not initially mark any passenger as boarded.
- The Python script, upon execution, need to produce a complete PDDL problem file that in :init, gives slow1 and slow2 capacity levels n1 through n4 and gives fast1 capacity levels n1 through n6 using the domain's can-hold predicate.
- The Python script, upon execution, need to produce a complete PDDL problem file that in :init, includes every adjacent next relation from n0 through n(F-1) and every required above relation from each lower floor object to every higher floor object, without reversed or self relations.
- The Python script, upon execution, need to produce a complete PDDL problem file that in :init, marks every declared floor as reachable by slow1, slow2, and fast1.
- The Python script, upon execution, need to produce a complete PDDL problem file that in :init, includes one travel-slow and one travel-fast numeric assignment for every lower-floor/higher-floor pair. For floors separated by d levels, the slow cost must be d + 5 and the fast cost must be 3d + 1.
- The Python script, upon execution, need to produce a complete PDDL problem file that initializes total-cost to 0 and includes the metric minimizing total-cost.
- The Python script, upon execution, need to produce a complete PDDL problem file that in :goal, assigns each of the 15 passengers to the destination floor stated for that passenger, using only declared objects and the domain's passenger-at predicate.
- The Python script, upon execution, need to produce a complete PDDL problem file that ensures parentheses are balanced and PDDL syntax is valid.


;; Example Second Stage Prompt
Review the existing first-stage Python generator against the original task prompt and fixed domain, then return a corrected and complete Python generator.

Concentrate on whether the program models each repeated structure with the correct source collection and loop bounds. The ordered floor/count values form one complete sequence from the first declared value through the last declared value. The generator must create every adjacent relationship across that entire sequence. Do not stop that sequence at the largest elevator capacity. Elevator capacity controls only how many passengers an elevator can hold, it must not be used as the bound for structures that range over all floors or all declared floor/count values.

Check all other generator invariants as well. Every elevator must be associated with every floor. Every lower-floor and higher-floor combination must be generated once, together with both travel costs computed from the difference between those floors. The slow-elevator cost is that difference plus five, and the fast-elevator cost is three times that difference plus one. Ensure that object collections, initial locations, elevator capacities, passenger destinations, and the conversion between human-numbered destinations and zero-based floor objects all come from the original description and use the correct bounds.

Correct the Python logic rather than manually expanding the resulting problem data. Keep repeated structures algorithmic and compact. Inspect loops for off-by-one errors, accidental use of one parameter in place of another, incomplete ranges, duplicate generation, inconsistent indexing, and formulas applied to the wrong pair of values. Preserve all parts of the draft that already agree with the task and fixed domain.

Return only the complete corrected Python source in exactly one <generator>...</generator> block, with no text before or after it. Do not return a problem file directly. Do not include Markdown fences, explanations, analysis, parser results, comparison results, expected fact lists, or manually supplied problem-file fragments. The generator must write the final problem file to problem.pddl in the current working directory.

Original task prompt:
{original_prompt}

Existing first-stage generator.py:
{draft_generator_code}

Problem file produced by the existing generator:
{draft_problem_pddl}
\end{lstlisting}

\begin{lstlisting}[language=lisp, caption={Prompt for LLM-as-Planner for Transport}, label={lst:prompt_planner_transport}]
    You are an expert automated planner. Your task is to read a natural-language domain description, the corresponding problem description, and the provided ground-truth domain PDDL, reason through the objects, actions, and goals, and output a valid plan.

Recall the distinctions:
- The domain description explains the general world: available action schemas, predicates, and object types.
- The problem description gives the concrete instance: specific objects, the initial state, and the goals for this planning episode.
- The ground-truth domain PDDL provides the formal acton schema and predicate definitions that your plan must follow exactly.

Planning output requirements:
- Produce the final plan inside <plan>...</plan>. Each action must be on its own line in classical PDDL plan syntax, e.g. `(move robot1 roomA roomB)`.
- Maintain the execution order from top to bottom. Do not include step numbers, timestamps, or probabilities.
- Only emit actions that are applicable given the initial state and that eventually achieve the goals.
- After the closing </plan> tag, do not add extra commentary.

Example format (replace names with the task-specific details):
<plan>
(pick-up truck-1 location-1 package-1 capacity-3 capacity-4)
(drive truck-1 location-1 location-2)
(drop truck-1 location-2 package-1 capacity-3 capacity-4)
</plan>

Follow this template for every problem. EOF
\end{lstlisting}

\begin{lstlisting}[language=lisp, caption={Prompt for LLM-as-Formalizer for Transport}, label={lst:prompt_formalizer_transport}]
    PDDL problem file contains problem name, domain name, objects in this problem instance, init state of objects, and goal state of objects.
Based on the natural language problem description, identify the relevant objects for this problem with their names and types.
Represent the initial state with the appropriate predicates and object arguments. Represent the goal state with the appropriate predicates and object arguments.
PDDL problem file has a definitive syntax that must be followed for any problem. An abstract example PDDL problem file is given below.

<problem_file>
(define
	(problem problem_name)
	(:domain domain_name)
	(:objects
		obj1 obj2 - type1
		obj3, obj4 - type2
	)
	(:init (predicate1 obj1 obj3) (predicate2 obj2 obj3))
	(:goal (and (predicate1 obj1 obj4) (predicate2 obj2 obj3)))
)
</problem_file>

Notes for generating problem file:
- obj1, obj2, ... are only representative and should be replaced with appropriate objects. There could be any number of objects with their types.
- init state with predicate1 & predicate2 is only representative and should be replaced with appropriate predicates that define init state
- goal state with predicate1 & predicate2 is only representative and should be replaced with appropriate predicates that define goal state
- predicates with proper arguments could be combined to combine complex boolean expression to represent init and goal states 
- The braces should be balanced for each section of the PDDL program
- Use predicates with arguments of the right type as declared in domain file
- All the objects that would be arguments of predicates in init and goal states should be declared in :objects


Additional strict constraints for large instances (10-30 locations):
- Output only one complete problem PDDL wrapped in <problem_file>...</problem_file>. Do not output any domain file.
- Include every required problem object once in :objects, and do not include undeclared or duplicate objects.
- Use only predicates and action-relevant symbols that are valid for the provided domain PDDL.
- In this Transport formulation, declare all locations, vehicles, packages, and capacity-number objects once using valid names and valid types from the problem description.
- In :init, include the full capacity-predecessor chain using only declared capacity-number objects and valid predicates.
- In :init, include the complete fully connected bidirectional road network between all distinct locations, and include the corresponding road-length facts exactly as required by the problem description.
- In :init, place each vehicle at one valid starting location and assign each vehicle the correct carrying capacity using only declared objects and valid predicates.
- In :init, place each package at one valid starting location, and do not place the same package at multiple locations or simultaneously inside a vehicle unless the problem description requires it.
- In :goal, include the required package destination facts for all and only the intended packages, using only declared objects and valid predicates, and avoid contradictory goals.
- Ensure parentheses are balanced and PDDL syntax is valid.

;; Example Second Stage Prompt
Review the first draft of the Transport problem file you just generated against the original task and the fixed Transport rules, then produce a corrected complete problem file.

A draft can appear plausible while quietly changing or omitting important instance details. Pay particular attention to the following:
- Preserve every name exactly as it appears in the original task, and keep each truck, package, location, and capacity level in its stated role. Do not replace names with shorter, longer, or more natural-sounding alternatives.
- Preserve the exact sequence of capacity levels described by the task and the precise carrying limit assigned to every truck. Do not omit a level, add an unsupported level, or infer different limits from truck numbering.
- Reproduce the complete directed travel network described by the task. When every location connects directly to every other distinct location, include every ordered direction rather than stopping early or summarizing the network.
- Preserve the stated distance for every directed connection. Treat travel from one location to another and travel in the reverse direction as separate details that must each be represented when required.
- Preserve the exact starting location of every truck and package. Do not move anything to a convenient default location or infer a different arrangement.
- Preserve each package's exact destination. Do not reorder destinations, shift assignments, or substitute names based on an intuitive pattern.
- When the task tracks travel cost, preserve its stated starting value and the instruction to prefer the lowest-cost solution.
- Do not introduce relationships, quantities, or conditions that are unsupported by the original task or the fixed rules. Keep every part of the draft that already agrees with the specification.

Treat the original task prompt as authoritative whenever it differs from the draft. Return the complete corrected problem file in one <problem_file>...</problem_file> block, with no text before or after it.

Original task prompt:
{original_prompt}
\end{lstlisting}

\begin{lstlisting}[language=lisp, caption={Prompt for LLM-as-Higher-Order-Formalizer for Transport}, label={lst:prompt_ho_formalizer_transport}]
;; Original Prompt
PDDL problem file contains problem name, domain name, objects in this problem instance, init state of objects, and goal state of objects.
Based on the natural language problem description, identify the relevant objects for this problem with their names and types.
In a formal PDDL problem file, initial state have the appropriate predicates and object arguments.
In a formal PDDL problem file, goal state also have the appropriate predicates and object arguments.
PDDL problem file has a definitive syntax that must be followed for any problem. An abstract example PDDL problem file is given below.

(define
	(problem problem_name)
	(:domain domain_name)
	(:objects
		obj1 obj2 - type1
		obj3, obj4 - type2
	)
	(:init (predicate1 obj1 obj3) (predicate2 obj2 obj3))
	(:goal (and (predicate1 obj1 obj4) (predicate2 obj2 obj3)))
)

You MUST generate a PYTHON SCRIPT that, upon execution, generates the PDDL problem file that matches the natural language problem description:
<generator>
the python script code...
</generator>

EXAMPLE executable generator style from another domain (generalized):

<generator>
def build_init_from_unraveling_rule(n):
  blocks = [f"block{i}" for i in range(1, n + 1)]

  # Odd and even stacks are TOP -> BOTTOM with increasing numbering
  odd_stack = [b for i, b in enumerate(blocks, start=1) if i % 2 == 1]
  even_stack = [b for i, b in enumerate(blocks, start=1) if i % 2 == 0]

  init_facts = []

  # Convert one stack (top->bottom) to on/on-table/clear facts
  def emit_stack(stack):
      if not stack:
          return
      for idx, block in enumerate(stack):
          if idx == len(stack) - 1:
              init_facts.append(f"(on-table {block})")
          else:
              under = stack[idx + 1]
              init_facts.append(f"(on {block} {under})")
          if idx == 0:
              init_facts.append(f"(clear {block})")

  emit_stack(odd_stack)
  emit_stack(even_stack)

  init_facts.append("(arm-empty)")
  return blocks, init_facts


def build_problem_pddl(problem_name, domain_name, blocks, init_facts, goal_facts):
  lines = []
  lines.append(f"(define (problem {problem_name})")
  lines.append(f"  (:domain {domain_name})")
  lines.append(f"  (:objects {' '.join(blocks)})")
  lines.append("  (:init")
  for fact in init_facts:
      lines.append(f"    {fact}")
  lines.append("  )")
  lines.append("  (:goal (and")
  for fact in goal_facts:
      lines.append(f"    {fact}")
  lines.append("  ))")
  lines.append(")")
  return "\n".join(lines) + "\n"


def main():
  # n should be inferred from the given NL problem description in the real task
  n = 10
  problem_name = "example_problem"
  domain_name = "xxx"

  blocks, init_facts = build_init_from_unraveling_rule(n)

  # Goal extraction can be abstracted; replace with parsed goal facts from the given NL problem
  goal_facts = [
      "(on block1 block5)",
      "(on block2 block1)",
      "(on-table block4)"
  ]

  pddl_text = build_problem_pddl(problem_name, domain_name, blocks, init_facts, goal_facts)

  with open("problem.pddl", "w", encoding="utf-8") as f:
      f.write(pddl_text)


if __name__ == "__main__":
  main()
</generator>


Notes for generating the python script:
- The Python script is ONLY TRANSLATING the natural language problem description to the formal PDDL problem file.
- The Python script is ABSOLUTELY NOT SOLVING the Transport problem


Additional strict constraints for all instances (10-30 locations):
- Output only one complete Python script wrapped in <generator>...</generator>. Do not output any domain file.
- The script MUST derive init/goal facts algorithmically from compact structure, not hardcoded full fact lists.
- Disallowed: manually listing full init\_state/goal\_state facts as constants.
- Required: at least one loop that constructs predicates (e.g., road/location/at/at-package/capacity) from computed patterns.
- Required: compute object lists from inferred location count, package count, and vehicle count (do not hardcode all object names manually).
- The Python script, upon execution, need to produce a complete PDDL problem file that includes every location, package, and vehicle mentioned in the problem description once in :objects and do not include undeclared objects.
- The Python script, upon execution, need to produce a complete PDDL problem file that uses only predicates and action-relevant symbols that are valid for the provided domain PDDL.
- The Python script, upon execution, need to produce a complete PDDL problem file that in :init, places each vehicle at one declared location and each package at one declared location.
- The Python script, upon execution, need to produce a complete PDDL problem file that in :init, assigns the fixed carrying capacity to every vehicle using only valid capacity-related symbols from the provided domain PDDL.
- The Python script, upon execution, need to produce a complete PDDL problem file that in :init, derives the bidirectional road connectivity facts algorithmically from the described fully connected transport network and avoids invalid self-connections unless they are explicitly described.
- The Python script, upon execution, need to produce a complete PDDL problem file that in :goal, assigns each package to one declared destination location, uses only declared objects and valid predicates, and avoids contradictory goals.
- The Python script, upon execution, need to produce a complete PDDL problem file that ensures parentheses are balanced and PDDL syntax is valid.

;; Example Second Stage Prompt
Review the existing first-stage Python generator and the problem file it produced against the original task prompt and fixed domain. Return a corrected, complete Python generator.

Pay special attention to the ordering of capacity levels. Each smaller capacity value must be treated as immediately preceding the next larger capacity value, beginning with zero and continuing one level at a time to the maximum, do not reverse this ordering. Explicitly assign every declared entity to its correct category: locations, packages, vehicles, and capacity values. Do not combine all declared entities into a single untyped list. Correct the generator itself so that these requirements are reflected throughout the problem file it produces, while preserving parts of the draft that already agree with the task and domain.

Return the complete corrected Python generator in one <generator>...</generator> block, with no text before or after it. The generator must write the final problem file to problem.pddl in the current working directory.

Original task prompt:
{original_prompt}

Existing first-stage generator.py:
{draft_generator_code}

Problem file produced by the existing generator:
{draft_problem_pddl}
\end{lstlisting}

\end{document}